\documentclass[pdflatex,sn-mathphys-num]{sn-jnl}% Math and Physical Sciences Numbered Reference Style
\usepackage{graphicx}%
\usepackage{multirow}%
\usepackage{amsmath,amssymb,amsfonts}%
\usepackage{amsthm}%
\usepackage{mathrsfs}%
\usepackage[title]{appendix}%
\usepackage{xcolor}%
\usepackage{textcomp}%
\usepackage{manyfoot}%
\usepackage{booktabs}%
\usepackage{algorithm}%
\usepackage{algorithmicx}%
\usepackage{algpseudocode}%
\usepackage{listings}%
\theoremstyle{thmstyleone}%
\theoremstyle{thmstyletwo}%

\theoremstyle{thmstylethree}%

\begin{document}

\title[C2L-ST]{Central-to-Local Adaptive Generative Diffusion Framework for Improving Gene Expression Prediction in Data-Limited Spatial Transcriptomics}

%%=============================================================%%
%% GivenName	-> \fnm{Joergen W.}
%% Particle	-> \spfx{van der} -> surname prefix
%% FamilyName	-> \sur{Ploeg}
%% Suffix	-> \sfx{IV}
%% \author*[1,2]{\fnm{Joergen W.} \spfx{van der} \sur{Ploeg} 
%%  \sfx{IV}}\email{iauthor@gmail.com}
%%=============================================================%%

% \author*[1]{\fnm{Yaoyu} \sur{Fang}}\email{yaoyu.fang@northwestern.edu}

% \author[1]{\fnm{Jiahe} \sur{Qian}}\email{qianjiahe2023@ia.ac.cn}
% % \equalcont{These authors contributed equally to this work.}

% \author[2]{\fnm{Xinkun} \sur{Wang}}\email{xinkun.wang@northwestern.edu}
% % \equalcont{These authors contributed equally to this work.}

% \author[3]{\fnm{Lee A.} \sur{Cooper}}\email{lee.cooper@northwestern.edu}

% \author*[1]{\fnm{Bo} \sur{Zhou}}\email{bo.zhou@northwestern.edu}

\author*[1]{\fnm{Yaoyu} \sur{Fang}}\email{yaoyu.fang@northwestern.edu}

\author[1]{\fnm{Jiahe} \sur{Qian}}
% \equalcont{These authors contributed equally to this work.}

\author[2]{\fnm{Xinkun} \sur{Wang}}
% \equalcont{These authors contributed equally to this work.}

\author[3]{\fnm{Lee A.} \sur{Cooper}}

\author*[1]{\fnm{Bo} \sur{Zhou}}\email{bo.zhou@northwestern.edu}

% \affil[1]{\orgdiv{Department of Radiology}, \orgname{Northwestern University}, \orgaddress{\street{676 N. St. Clair St.}, \city{Chicago}, \postcode{60611}, \state{IL}, \country{USA}}}

% \affil[2]{\orgdiv{Department of Cell and Developmental Biology}, \orgname{Northwestern University}, \orgaddress{\street{303 E. Superior St.}, \city{Chicago}, \postcode{60611}, \state{IL}, \country{USA}}}

% \affil[3]{\orgdiv{Department of Pathology}, \orgname{Northwestern University}, \orgaddress{\street{303 East Chicago Avenue}, \city{Chicago}, \postcode{60611}, \state{IL}, \country{USA}}}

\affil[1]{\orgdiv{Department of Radiology}, \orgname{Northwestern University}}

\affil[2]{\orgdiv{Department of Cell and Developmental Biology}, \orgname{Northwestern University}}

\affil[3]{\orgdiv{Department of Pathology}, \orgname{Northwestern University}}

%%==================================%%
%% Sample for unstructured abstract %%
%%==================================%%

\abstract{Spatial transcriptomics (ST) provides spatially resolved gene expression profiles within intact tissue architecture, enabling molecular analysis in histological context. However, the high cost, limited throughput, and restricted data sharing of ST experiments result in severe data scarcity, constraining the development of robust computational models. To address this limitation, we present a Central-to-Local adaptive generative diffusion framework for ST (C2L-ST) that integrates large-scale morphological priors with limited molecular guidance. A global central model is first pretrained on extensive histopathology datasets to learn transferable morphological representations, and institution-specific local models are then adapted through lightweight gene-conditioned modulation using a small number of paired image–gene spots. This strategy enables the synthesis of realistic and molecularly consistent histology patches under data-limited conditions. The generated images exhibit high visual and structural fidelity, reproduce cellular composition, and show strong embedding overlap with real data across multiple organs, reflecting both realism and diversity. When incorporated into downstream training, synthetic image–gene pairs improve gene expression prediction accuracy and spatial coherence, achieving performance comparable to real data while requiring only a fraction of sampled spots. C2L-ST provides a scalable and data-efficient framework for molecular-level data augmentation, offering a domain-adaptive and generalizable approach for integrating histology and transcriptomics in spatial biology and related fields.}

\keywords{Spatial transcriptomics, Generative AI, Diffusion model, Central-to-local adaptation, Gene expression prediction}

%%\pacs[JEL Classification]{D8, H51}

%%\pacs[MSC Classification]{35A01, 65L10, 65L12, 65L20, 65L70}

\maketitle

\section{Introduction}\label{sec1}

% \cite{bib1}

Spatial transcriptomics (ST) enables the spatially resolved mapping of gene expression within intact tissue architecture, linking molecular profiles to histological context~\cite{andersonSpatialTranscriptomics2022,aspSpatiallyResolvedTranscriptomes2020}. ST provides a robust foundation for elucidating disease mechanisms, identifying spatial heterogeneity within tumors, and guiding precision diagnostics and therapeutic decision-making~\cite{anSpatialTranscriptomicsBreast2024,aungSpatiallyInformedGene2024,burgessSpatialTranscriptomicsComing2019}. Recent improvements in sequencing chemistry, capture strategies, and computational analysis have markedly enhanced resolution and throughput, establishing ST as a powerful approach for decoding complex tissue microenvironments and cellular interactions~\cite{boeSpatialTranscriptomicsReveals2024,attaComputationalChallengesOpportunities2021}. Unlike conventional transcriptomic techniques, such as single-cell RNA sequencing that generates detailed molecular profiles but loses positional information, ST retains the spatial context essential for discovering disease biomarkers and understanding pathological progression~\cite{aungSpatiallyInformedGene2024,boeSpatialTranscriptomicsReveals2024,sankar50480ImmuneInfiltrate2024}. 

Despite recent advances, ST data remain costly and limited in availability, making it challenging to obtain sufficient molecular measurements at high spatial resolution~\cite{attaComputationalChallengesOpportunities2021}. This limitation has motivated the development of methods for predicting gene expression directly from histology, a task that is particularly important because it not only provides insights into cellular functions and disease mechanisms but also provides a cost-efficient alternative. However, despite the proposal of many models such as ST-Net~\cite{pangLeveragingInformationSpatial2021}, EGN~\cite{yangExemplarGuidedDeep2023}, EGGN~\cite{yangSpatialTranscriptomicsAnalysis2024c}, HGGEP~\cite{liGeneExpressionPrediction2024a} and DANet~\cite{wuDANetSpatialGene2025}, progress in histology-based gene expression prediction remains constrained by multiple factors. Publicly accessible ST datasets are still scarce and unevenly distributed. Generating high-resolution data requires expensive reagents, sophisticated equipment, and specialized labor, making large-scale studies financially prohibitive~\cite{fangComputationalApproachesChallenges2023,smithChallengesOpportunitiesClinical2024}. 
%Moreover, variations across samples and sequencing platforms complicate integration and reduce model robustness. Data sharing is further restricted by strict privacy and regulatory constraints: pathology images are highly sensitive, and many institutions are unable to release patient-derived molecular profiles due to confidentiality concerns~\cite{padariyaPrivacyPreservingGenerativeModels2025,xinArtificialIntelligenceCentral2024}. Even when datasets exist, regulatory compliance often results in proprietary control or limited accessibility~\cite{choeAdvancesChallengesSpatial2023,fangComputationalApproachesChallenges2023}. 
Moreover, variations across samples and sequencing platforms complicate integration and reduce model robustness. Data sharing is further restricted by institutional policies and proprietary controls~\cite{padariyaPrivacyPreservingGenerativeModels2025,xinArtificialIntelligenceCentral2024}. Even when datasets exist, regulatory compliance and data ownership often lead to limited accessibility or usage restrictions~\cite{choeAdvancesChallengesSpatial2023,fangComputationalApproachesChallenges2023}.
In addition, dataset imbalance also poses significant challenges for training deep learning models. Together, these technical, economic, and regulatory barriers result in a shortage of diverse, open, and high-quality ST datasets, which continue to limit the effectiveness, scalability, and generalizability of current gene expression prediction approaches.

% Generative models offer a promising means to alleviate the data scarcity barriers described above. 
Generative models provide a promising solution to address data scarcity by synthesizing realistic samples that expand training distributions without requiring additional experimental data. In digital histopathology, diffusion models synthesize realistic tissue images and support data augmentation, enabling robust downstream analysis while reducing reliance on expensive, annotated datasets; these models exhibit greater stability in image generation, cover richer distributions, and demonstrate enhanced resistance to noise, making them particularly suitable for pathology images that often feature significant staining variations and artifacts. Recent work has further extended these advantages by conditioning diffusion models on different sources of information, which can be grouped into three categories: textual information, morphological information, and gene expression data~\cite{yellapragadaPixCellGenerativeFoundation2025}. 
Text-conditioned diffusion models leverage natural language or pathology reports as guiding signals. Rao et al. (2024) refined text-driven latent diffusion for cancer pathology, improving the alignment between textual prompts and generated images, while Yellapragada et al. introduced PathLDM, the first model to jointly condition on pathology images and GPT-derived pathology report summaries, achieving state-of-the-art fidelity scores and validating the use of text-guided synthesis for downstream classification~\cite{raoImprovingTextconditionedLatent2024, yellapragadaPathLDMTextConditioned2023}.
Morphology-guided approaches incorporate structural, cellular, or multimodal image-based cues. ToPoFM synthesizes cell arrangements with large language models and uses diffusion models to generate high-resolution pathology images with cell-level control, improving classification and segmentation tasks~\cite{liToPoFMTopologyGuidedPathology2025}. Lou et al. integrated hematoxylin and eosin (H\&E) and immunohistochemistry (IHC) modalities in a multimodal denoising diffusion pretraining framework, enriching whole slide images (WSI) classifiers with cross-stain information~\cite{louMultimodalDenoisingDiffusion2024}. Niehues et al. evaluated latent diffusion models as privacy-preserving augmenters, showing that synthetic colorectal cancer images can enhance classification while mitigating data sharing risks~\cite{niehuesUsingHistopathologyLatent2024}. PixCell, the generative foundation model for histopathology, trained on tens of millions of tiles, produces high-fidelity synthetic images across cancer types, supports virtual staining, and serves as a data substitute for both segmentation and representation learning~\cite{yellapragadaPixCellGenerativeFoundation2025}. At a finer granularity, Oh et al. introduced DiffMix to augment imbalanced nuclei datasets for segmentation and classification~\cite{ohDiffMixDiffusionModelbased2023}, while Shrivastava et al. developed NASDM, a nuclei-aware semantic generator that directly models nuclear morphology for biologically faithful synthesis~\cite{shrivastavaNASDMNucleiAwareSemantic2023}. Xu et al. proposed ViT-DAE, a Transformer-driven diffusion autoencoder generating diverse, high-quality images and improving small-scale dataset augmentation~\cite{xuViTDAETransformerdrivenDiffusion2023}. 
However, the above approaches do not integrate histology with gene expression and therefore remain at the morphological level without reaching molecular resolution. Gene expression–guided models represent a more recent direction that connects molecular data with image synthesis. RNA-CDM is a cascaded diffusion framework that conditions on bulk RNA-seq profiles to generate synthetic H\&E tiles. By encoding high-dimensional gene expression into a latent space with a $\beta$-VAE and guiding diffusion through this representation, RNA-CDM produces histology-consistent images that preserve tumor–immune cell composition and serve as valuable synthetic data for downstream classification tasks~\cite{carrillo-perezGenerationSyntheticWholeslide2025a}. Nevertheless, because it relies on bulk RNA-seq, the approach still lacks spatial and single-cell resolution, limiting its capacity to fully capture fine-grained gene expression.

Despite progress in diffusion-based modeling, applying generative models directly to ST remains challenging~\cite{chenOpportunitiesChallengesDiffusion2024}. Model training often requires vast amounts of data, whereas ST datasets are inherently limited by high experimental costs, labor-intensive protocols, and technical variability across platforms~\cite{alajajiGenerativeAdversarialNetworks2024}. In addition, the sharing of real patient-derived transcriptomic and pathology data is heavily restricted by privacy and regulatory constraints, further exacerbating the scarcity of accessible ST resources~\cite{moghadamMorphologyFocusedDiffusion2022,niehuesUsingHistopathologyLatent2024}. Moreover, most existing generative frameworks are tailored to specific scenarios, making them difficult to generalize across tissue types or experimental conditions, while also incurring computational and training costs~\cite{ohDiffMixDiffusionModelbased2023,carrillo-perezGenerationSyntheticWholeslide2025a,liToPoFMTopologyGuidedPathology2025}. In terms of downstream applications, current models are still primarily evaluated on auxiliary tasks such as image classification or nuclei segmentation, and therefore contribute little to molecular-level discovery, such as gene expression prediction from histology, which is a critical task in ST, as they rarely incorporate spatially resolved expression profiles as conditioning signals.

\begin{figure*}[htbp!]
    \centering
    \includegraphics[width=0.98\textwidth]{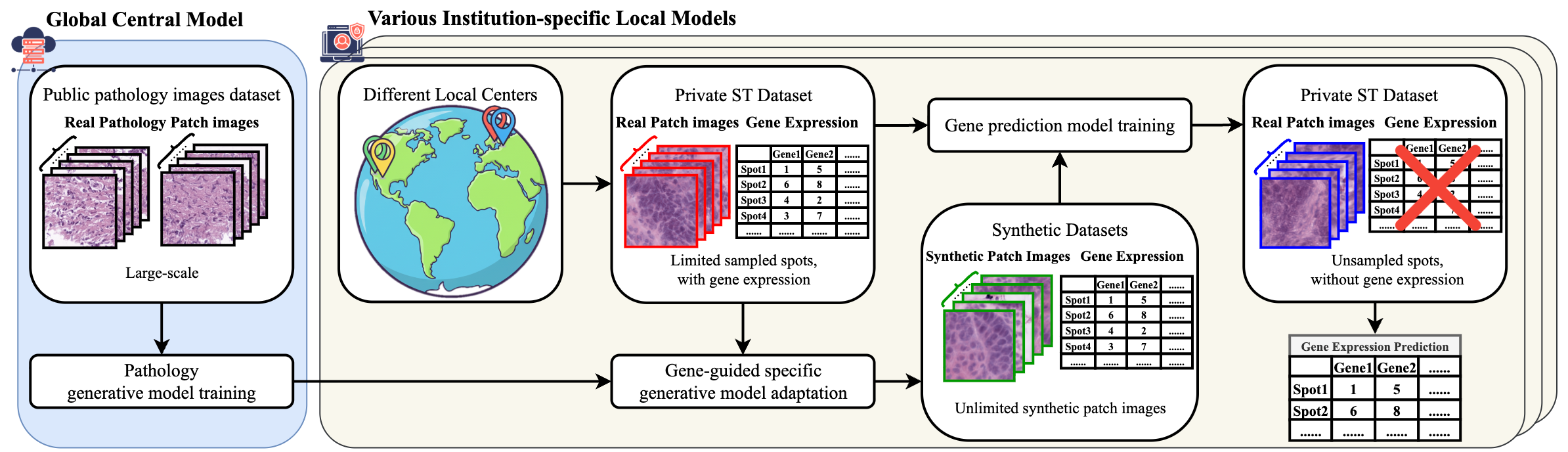}
    \caption{\textbf{Overview of the C2L-ST framework for central-to-local adaptive generative modeling in spatial transcriptomics.}  The framework integrates large-scale public histopathology data with limited local ST data to enable molecular-level generation across centers worldwide. The central generative diffusion model is first pretrained on large-scale histopathology image patches to learn diverse tissue morphological priors. Institution-specific local models from different centers are then adapted using a small number of paired image–gene samples (red), aligning morphological features with molecular information to produce synthetic histology patches conditioned on gene expression (green). The resulting synthetic data expand the effective training distribution without requiring the exchange of real patient data across institutions. The adapted local models are further applied to unsampled tissue regions (blue) to predict gene expression directly from histopathology images. This central-to-local strategy alleviates data scarcity by generating synthetic samples, reduces domain discrepancies through shared morphological priors, and respects institutional data privacy by allowing each center to adapt models locally without data exchange.}
    \label{fig:framework}
\end{figure*}

To address these limitations, we propose a central-to-local generative framework for ST (C2L-ST), as illustrated in Fig.~\ref{fig:framework}. In the first stage, we establish a global central model by pre-training an unconditional diffusion model on large-scale histopathology datasets. This model captures diverse morphological patterns and provides a transferable morphological prior. Importantly, the central model is designed with the flexibility to incorporate additional conditioning signals during adaptation, enabling extension beyond unconditional generation. In the second stage, institution-specific local models are derived from the central model. If an institution requires support for ST, the central model is delivered and adapted through lightweight conditional modulation using the limited paired image–gene expression spots available at that site. Through this gene-conditioned adaptation, the local model produces institution-specific synthetic data that augment scarce real samples and directly support downstream tasks, including improving the accuracy of gene expression prediction at unsampled spatial locations. Because adaptation and data usage occur entirely within each institution, sensitive patient-level data never need to be shared externally, ensuring that privacy is preserved while still benefiting from the shared global prior. We validate our approach across multiple tissue types and independent local adaptation tests, demonstrating that the generated images exhibit high visual quality and that the locally adapted models consistently improve downstream gene expression prediction. This framework combines the scalability of a shared global generative prior with the flexibility of local adaptation, providing a data-efficient and generalizable approach for integrating morphological and molecular information in ST.
% This framework combines the scalability of a shared global generative prior with the flexibility of local adaptation, offering a data-efficient and privacy-aware pathway for integrating morphology and molecular signals in ST.

\section{Methods}

\begin{figure*}[htbp!]
    \centering
    \includegraphics[width=0.95\textwidth]{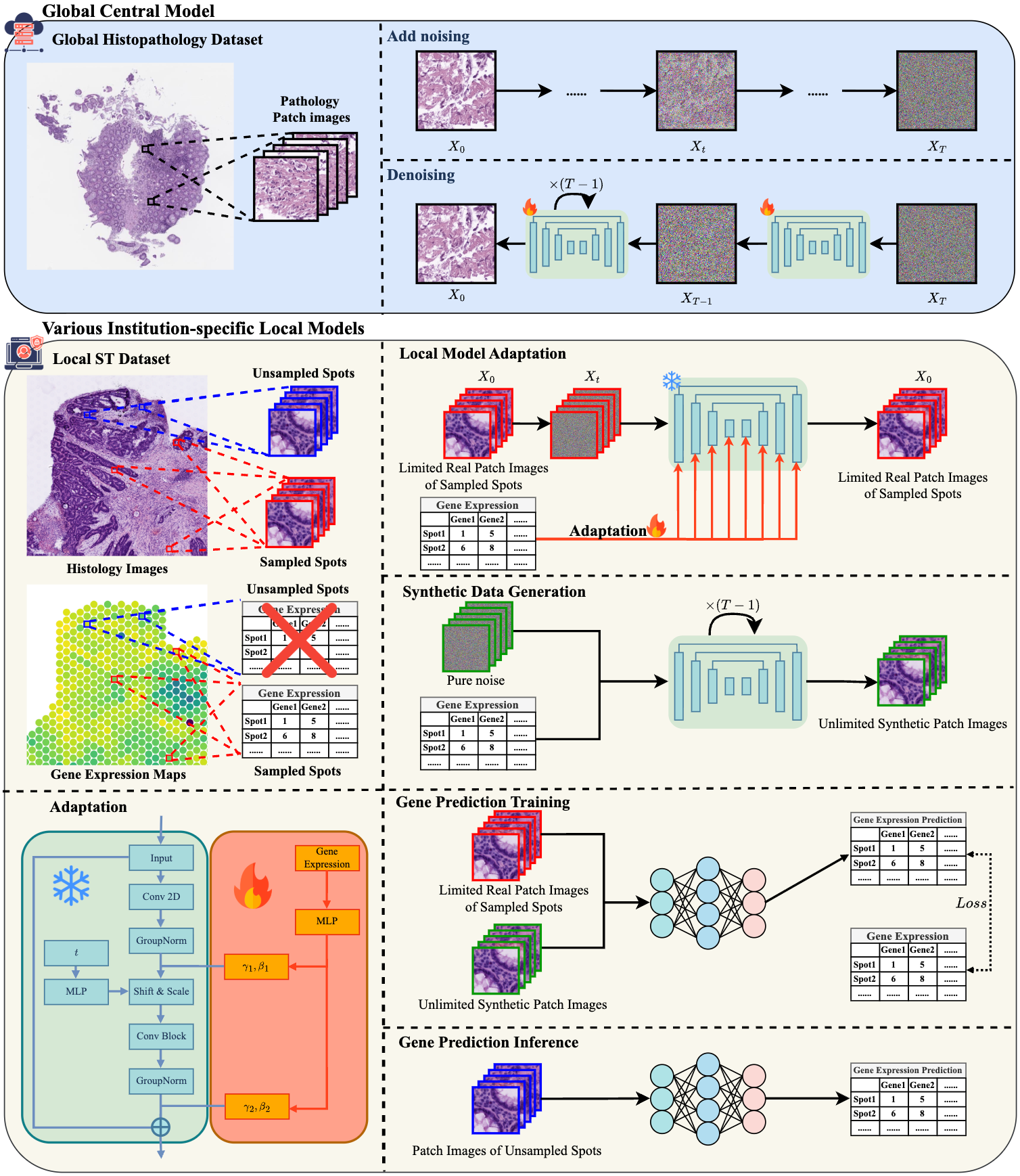}
    \caption{\textbf{Detailed workflow of the C2L-ST framework integrating central pretraining and local adaptation.} The central generative diffusion model is first trained on large-scale public histopathology datasets to learn transferable tissue morphological priors through iterative noising and denoising processes. Each institution/center then performs local adaptation using its private ST dataset, where histology patches and corresponding gene expression values (red) guide lightweight adaptation of the pretrained model to align morphology with molecular information. After adaptation, the local model generates synthetic histology patches conditioned on gene expression (green), expanding the effective dataset. These synthetic samples are combined with limited real data to train gene prediction networks that learn the mapping from histopathology images to gene expression. The trained predictors are subsequently applied to unsampled tissue regions (blue) to infer gene expression profiles directly from histology images. This unified pipeline enables molecular-level data augmentation, improves prediction performance under data-limited conditions, and maintains data confidentiality across centers.}
    \label{fig:details}
\end{figure*}

We propose a generative framework for ST that alleviates data scarcity and institutional data-sharing constraints by integrating large-scale histopathology resources with limited spatially resolved gene expression measurements. The framework proceeds in two stages: (i) establishing a global central model by pre-training an unconditional diffusion model on abundant histopathology image collections to learn diverse morphological representations, and (ii) adapting this central model into institution-specific local models through lightweight gene-conditioned modulation using only a few paired image–gene expression spots. The adapted local models generate synthetic image–gene expression pairs that are combined with real data to enhance downstream gene expression prediction at unsampled spatial locations. The overall framework of our proposed method is illustrated in Fig.~\ref{fig:framework}. This design enables scalable and cost-efficient integration of morphological and molecular signals while maintaining institutional data governance by allowing each center to adapt the central model locally without sharing raw data.

\subsection{Global central model pre-training for morphological prior learning}\label{subsec:diff_training}

The first stage of our framework focuses on constructing a global central model by pre-training an unconditional diffusion model on large-scale publicly available histopathology image datasets (Fig.~\ref{fig:details}). The objective is to capture diverse morphological structures and establish a transferable generative prior that can later be adapted to institution-specific datasets, while requiring no access to paired molecular measurements. 

The diffusion process is fundamentally defined by two probabilistic models: the forward process $q$, which consists of a fixed, parameterized chain of noising distributions $q(x_t|x_{t-1})$; and the reverse process $p$, which is a denoising distribution $p_\theta(x_{t-1}|x_t)$ learned by the model to generate data. In the forward process, an image $X^H_0$ sampled from the data distribution $p_{\text{data}}(X^H)$ is gradually perturbed by Gaussian noise according to a predefined variance schedule $\{\beta_t\}_{t=1}^T$:

\begin{align}
q(X^H_t \mid X^H_{t-1}) &= \mathcal{N}\!\left(X^H_t; \sqrt{1-\beta_t}\, X^H_{t-1}, \, \beta_t I \right), \\
q(X^H_{1:T} \mid X^H_0) &= \prod_{t=1}^T q(X^H_t \mid X^H_{t-1}),
\end{align}

where $\beta_t \in (0,1)$ controls the noise variance at step $t$. We also define $\alpha_t = 1-\beta_t$ and the cumulative product $\bar{\alpha}_t = \prod_{s=1}^t \alpha_s$, which allows direct sampling of noisy images at arbitrary step $t$:

\begin{align}
q(X^H_t \mid X^H_0) = \mathcal{N}\!\left(X^H_t; \sqrt{\bar{\alpha}_t} \, X^H_0, \, (1-\bar{\alpha}_t) I \right).
\end{align}

In the reverse process, the goal is to learn a parameterized distribution $p_\theta$ that approximates the reverse Markov chain that reconstructs clean images from noise:

\begin{align}
p_\theta(X^H_{0:T}) &= p(X^H_T) \prod_{t=1}^T p_\theta(X^H_{t-1} \mid X^H_t), \\
p_\theta(X^H_{t-1} \mid X^H_t) &= \mathcal{N}\!\left(X^H_{t-1}; \mu_\theta(X^H_t, t), \, \Sigma_\theta(X^H_t, t)\right).
\end{align}

Directly parameterizing $\mu_\theta$ and $\Sigma_\theta$ is intractable. Following the Denoising Diffusion Probabilistic Models (DDPM)~\cite{hoDenoisingDiffusionProbabilistic2020} framework, we instead train a neural network $\epsilon_\theta$ to predict the noise component $\epsilon \sim \mathcal{N}(0,I)$ that was added during the forward process. By accurately estimating this noise, the network implicitly learns the reverse dynamics needed to recover $X^H_0$ from $X^H_t$. The training objective reduces to:
\begin{align}
\mathcal{L}_{\text{diff}}(\theta) = \mathbb{E}_{X^H_0, \epsilon, t} 
\left[ \, \|\epsilon - \epsilon_\theta(X^H_t, t)\|^2 \, \right],
\end{align}
where $\epsilon \sim \mathcal{N}(0, I)$ and $X^H_t$ are generated according to $q(X^H_t \mid X^H_0)$. 
This objective defines the learning problem as fitting a generative model $p_\theta$ that approximates the true data distribution $p_{\text{data}}(X^H)$ by reversing the noising process. To parameterize $\epsilon_\theta$, we adopt a U-Net architecture, which has been widely used in diffusion models for image synthesis. Once trained, the diffusion model generates synthetic pathology images by starting from random Gaussian noise $X^H_T \sim \mathcal{N}(0,I)$ and iteratively applying the learned reverse transitions,
\begin{align}
X^H_{t-1} = \mu_\theta(X^H_t, t) + \Sigma_\theta(X^H_t, t)^{1/2} z, \quad z \sim \mathcal{N}(0, I),
\end{align}
until reaching $X^H_0$, which serves as a synthetic sample drawn from the learned distribution.

\subsection{Institution-specific local model adaptation with gene-conditioned modulation}\label{subsec:fine_tuning}

To adapt the global central model to ST, we develop institution-specific local models through lightweight gene-conditioned modulation on datasets where paired histology images and gene expression measurements are available. Specifically, each sampled spot provides both a histology patch $X^H$ and a corresponding gene expression profile $G \in \mathbb{R}^d$, where $d$ denotes the number of measured genes. These paired observations guide the conditional adaptation process, enabling the model to capture dependencies between morphological patterns and molecular signals.

We formulate the adaptation of the central model as a conditional diffusion problem, where the reverse process is guided by $G$. At each denoising step, the conditional distribution is defined as

\begin{align}
p_\theta(X^H_{t-1} \mid X^H_t, G) 
= \mathcal{N}\!\left(X^H_{t-1}; \mu_\theta(X^H_t, t, G), \, \Sigma_\theta(X^H_t, t, G)\right).
\end{align}

To inject the conditioning signal, the high-dimensional gene expression vector $G$ is first compressed into a low-dimensional latent representation using a learnable projection layer. This projection is trained jointly with the adaptation process, ensuring that the compressed embedding captures the most informative features of $G$. From this embedding, a set of scaling and shifting coefficients $(\gamma, \beta)$ is derived and applied to the normalization layers within specific ResNet~\cite{heDeepResidualLearning2015} blocks of the U-Net backbone:
\begin{align}
h' = \gamma(G) \cdot h + \beta(G),
\end{align}
where $h$ denotes the intermediate feature maps and $(\gamma(G), \beta(G))$ are the modulation factors generated from the projected gene embedding. During adaptation, all original U-Net parameters are frozen, and only the projection layer and the mapping from $G$ to $(\gamma, \beta)$ are updated. 

The conditional denoising objective is then defined as

\begin{align}
\mathcal{L}_{\text{cond}}(\theta) = 
\mathbb{E}_{X^H_0, G, \epsilon, t} 
\left[ \, \|\epsilon - \epsilon_\theta(X^H_t, t, G)\|^2 \, \right],
\end{align}

where $\epsilon \sim \mathcal{N}(0,I)$ and $X^H_t$ are generated from the forward process conditioned on the observed $X^H_0$. 

By adapting the pretrained diffusion model with only a small number of paired spots, the network learns to align morphological features with gene expression signals. This step enables the generation of synthetic image–gene expression pairs that preserve both histological realism and molecular consistency, thereby connecting the domains of pathology images with transcriptomic data. Moreover, since the generated image–gene pairs are synthetic and do not correspond to any individual patient, they offer a de-identified and governance-compliant alternative to sensitive ST measurements, enabling safer data sharing and broader use in downstream analyzes.

\subsection{Gene expression prediction via co-training with real and synthetic pairs from local adaptation}\label{subsec:cotraining}

After local adaptation, the conditional diffusion model can generate synthetic image–gene expression pairs $(X^H, G)$ that mimic real observations while remaining free of identifiable patient information. These synthetic pairs are combined with the limited real ST samples in a co-training strategy, enabling the predictive model to leverage both sources of data for more accurate gene expression inference at unsampled spatial locations. 

Formally, let $\mathcal{D}_{\text{real}} = \{(X^H_i, G_i)\}_{i=1}^{N}$ denote the set of sampled real spots from the target ST dataset, and $\mathcal{D}_{\text{syn}} = \{(\tilde{X}^H_j, \tilde{G}_j)\}_{j=1}^{M}$ denote the synthetic pairs generated by the fine-tuned diffusion model. The two datasets are combined into a co-training set. 
\[
\mathcal{D}_{\text{train}} = \mathcal{D}_{\text{real}} \cup \mathcal{D}_{\text{syn}}.
\]
A downstream predictor $f_\phi$ is then optimized on $\mathcal{D}_{\text{train}}$ to map histology images to gene expression estimates:
\begin{align}
\mathcal{L}_{\text{pred}}(\phi) = 
\mathbb{E}_{(X^H, G) \sim \mathcal{D}_{\text{train}}} 
\left[ \, \| G - f_\phi(X^H)\|_1 \, \right].
\end{align}

This co-training strategy enables the predictor to exploit the diversity of synthetic samples while being grounded in real paired measurements. 

% \subsection{multi-dataset set up}\label{subsec:implement}
% implement details

\subsection{Central and local dataset setup}\label{subsec:datasets}

\textbf{Central dataset for global model pretraining:} Four global central models were individually pretrained on organ-specific subsets from the publicly available SPIDER dataset released by HistAI~\cite{nechaevSPIDERComprehensiveMultiOrgan2025}, one of the largest curated histopathology resources providing high-resolution, expert-annotated whole-slide images across multiple tissue types. Specifically, models were trained on skin, colorectal, thoracic, and breast tissues to ensure broad morphological coverage and enhance generalization. The central models were trained using 159{,}854 patches for skin, 77{,}182 for colorectal, 78{,}307 for thoracic, and 92{,}892 for breast, totaling over 400{,}000 image patches. This large-scale, organ-specific dataset provides a diverse morphological foundation for learning transferable structural representations applicable to various ST domains.
All patches were stain-normalized using the \texttt{staintools} package, where each organ was aligned to a representative reference patch to establish a unified stain space and reduce inter-sample variability, ensuring consistent color appearance across organs~\cite{zhouCielAlTorchstaintoolsV1042024}.

\textbf{Local datasets for adaptation and evaluation:} To evaluate and adapt the global model, we utilized four real-world ST datasets obtained from independent institutions, each representing a distinct tissue type and experimental condition. \textbf{\textit{Center 1}}: The \textit{bowel dataset} (ZEN48) was produced by SciLifeLab and KTH Royal Institute of Technology (Sweden) using the RNA-Rescue Spatial Transcriptomics (RRST) protocol~\cite{mirzazadehSpatiallyResolvedTranscriptomic2023}. Fresh-frozen human colon and ileum tissues collected at University Hospitals Leuven (Belgium) were analyzed to assess the recovery of spatial transcriptomic signals from partially degraded RNA samples. \textbf{\textit{Center 2}}: The \textit{breast dataset} (TENX13) was obtained from the 10$\times$~Genomics public repository, derived from an invasive ductal carcinoma (IDC) breast tissue block provided by BioIVT Asterand. The specimen corresponds to an AJCC/UICC stage~IIA tumor prepared as 10~$\mu$m cryosections. \textbf{\textit{Center 3}}: The \textit{skin dataset} (MEND40) originates from Stanford University School of Medicine~\cite{jiMultimodalAnalysisComposition2020} and contains ST and histological data from human squamous cell carcinoma (SCC) samples, designed to characterize tumor microenvironmental composition and spatial organization. \textbf{\textit{Center 4}}: The \textit{lung dataset} (MEND90), also generated by SciLifeLab and KTH (Sweden), represents healthy lung tissue analyzed with the RRST workflow~\cite{mirzazadehSpatiallyResolvedTranscriptomic2023}, serving as a control for assessing RNA integrity recovery in long-term frozen samples. In all datasets, each spatial spot corresponds to a $112\times112~\mu$m histology patch, forming paired image–gene expression samples used for local adaptation and downstream evaluation. Collectively, these four multi-center datasets encompass a broad spectrum of tissue states (i.e., healthy, diseased, and degraded), and reflect diverse institutional and experimental conditions, enabling a comprehensive and rigorous evaluation of the proposed central–local generative framework across both biological and technical variability. All local histology patches were stain-normalized using the same reference-based procedure as the central model training data, ensuring a consistent stain space between global and local domains.

\subsection{Implementation details}\label{subsec:implement}
All models were implemented in PyTorch and trained on NVIDIA A6000 GPUs. 

\textbf{Global central model pretraining:} For the diffusion backbone, we adopted a U-Net architecture with an initial channel dimension of 64 and four hierarchical feature scales $(1, 2, 4, 8)$. The model was optimized for 200,000 steps with a batch size of 256, gradient accumulation over 4 steps, and the AdamW optimizer (learning rate $1\times10^{-4}$, weight decay $1\times10^{-4}$). Exponential moving average (EMA) weights were updated every iteration with a decay factor of 0.999. The objective followed the $L_2$ noise-prediction loss with a minimum signal-to-noise ratio (SNR) weighting of 5. Conditional dropout ($p=0.1$) was applied to improve robustness.

\textbf{Local specific model adaptation:} For institution-specific adaptation, we used the pretrained diffusion backbone and introduced gene-conditioned modulation within the ResNet blocks while keeping all original parameters frozen. The conditioning vectors were derived from gene expression embeddings, projected through a one-layer MLP to produce adaptive $(\gamma, \beta)$ parameters for each selected ResNet block. Gene expression values were preprocessed using a $log(1+x)$ transformation and normalized. Each local adaptation used only a subset (typically 25\%) of spatial spots with paired image–gene expression data. The model was trained for 10,000 steps with a batch size of 32 and gradient accumulation over 8 steps. The learning rate was set to $1\times10^{-3}$ with AdamW optimization (weight decay $1\times10^{-4}$), EMA decay of 0.999, and linear warm-up over the first 3\% of total steps. During inference, image synthesis was carried out using the Denoising Diffusion Implicit Model (DDIM)~\cite{songDenoisingDiffusionImplicit2022} sampler with 250 timesteps. A conditioning scale of 3 was applied to strengthen the correspondence between gene expression inputs and generated histological structures while maintaining visual realism.

\subsection{Study designs and task evaluations}\label{subsec:evaluation}

\textbf{Data-limited evaluation design.} 
To assess the proposed framework under conditions that reflect realistic ST constraints, we adopted a data-limited evaluation protocol. In practice, each participating center or laboratory typically has access to only a few available ST sections due to the high experimental costs, limited sequencing throughput, and sample availability. Moreover, cross-center data sharing is often restricted by privacy regulations and institutional policies, while technical and biological variations across sites introduce domain gaps in staining, imaging, and tissue processing~\cite{attaComputationalChallengesOpportunities2021}. To reproduce and simulate these practical challenges, each local center in our evaluation contributed a single representative ST slide for model adaptation. 
Given the limited number of spatially resolved spots available in most current ST platforms, we additionally simulated sparse sampling conditions to create controlled evaluation settings with known ground truth, allowing systematic analysis of model adaptation under extreme data scarcity. Accordingly, only a small fraction of spatial spots (5\%, 10\%, 25\%, or 50\%) was randomly selected for local adaptation, with the remaining spots reserved for testing. This two-level restriction, combining limited section availability across centers with sparse spot coverage within sections, faithfully reproduces the practical constraints of current ST studies, ensuring that model evaluation aligns with real-world data conditions, even under extremely data-limited conditions. 

\textbf{Evaluation of generative performance.}  
The quality and biological consistency of the generated histology patches were assessed through both qualitative and quantitative analyses. Visually, generated images were compared with corresponding real histology patches conditioned on the same gene expression profiles. At the feature level, embeddings extracted from the Conch foundation model were used to compute distribution alignment via t-SNE visualization and feature-space distance metrics, evaluating how closely synthetic samples approximate the real data manifold~\cite{luVisuallanguageFoundationModel2024}. Additionally, cell segmentation and classification were performed using HoverNet~\cite{grahamHoVerNetSimultaneousSegmentation2019}, pretrained on the PanNuke dataset~\cite{gamperPanNukeDatasetExtension2020}, to quantify cellular composition and assess whether the generative model preserved realistic cell-type distributions.

\textbf{Evaluation of gene expression prediction.}  
For downstream gene expression prediction, without loss of generality, a ResNet-50~\cite{heDeepResidualLearning2015} architecture was employed as the regression backbone.  
The model was trained for 200 epochs with a cosine learning rate schedule starting at $1\times10^{-3}$ and decaying to a minimum of $1\times10^{-5}$, including a 5\% warm-up phase.  
The optimization objective minimized the mean absolute error (MAE) between predicted and ground-truth gene expression values.  
To ensure stable regression and biological relevance, the top 250 most highly expressed genes in each local center were selected as prediction targets.  
Models trained with both real and synthetic image–gene pairs were compared against those trained on real data alone to quantify the contribution of synthetic augmentation to predictive performance.

\section{Results}\label{sec2}

\subsection{Local gene-conditioned generation preserves morphology–molecular correspondence}

\begin{figure*}[htbp!]
    \centering
    \includegraphics[width=0.7\textwidth]{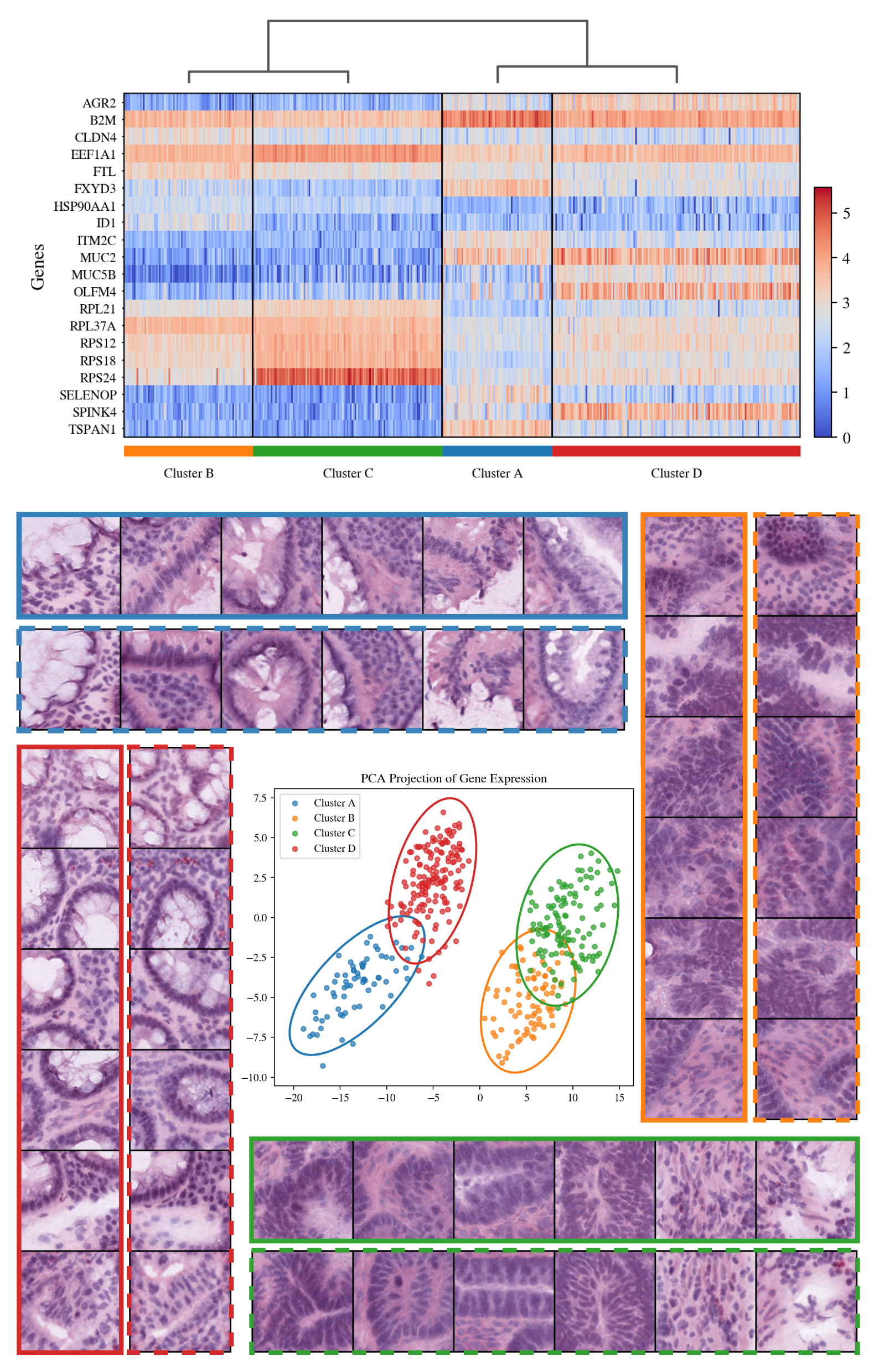}
    \caption{\textbf{Molecular and morphological correspondence in generated histology patches for the bowel (ZEN48) dataset.} The heatmap shows hierarchical clustering of representative genes, revealing four distinct molecular clusters (A–D). Each cluster is associated with characteristic histology patches from synthetic images, shown below with color-coded borders corresponding to the cluster identity. Dashed borders indicate synthetic images, and solid borders indicate real images. The PCA projection of gene expression highlights the separation among clusters in molecular space. Representative synthetic patches within each cluster display distinct structures consistent with their molecular signatures, indicating that the model captures the relationship between transcriptional programs and local morphology.}
    \label{fig:gen-ZEN48}
\end{figure*}

We assessed the visual and biological consistency of the locally adapted generative model by conditioning image synthesis on gene expression profiles from each ST center. As shown in Fig.~\ref{fig:gen-ZEN48}, clustering of gene expression in the bowel (ZEN48) center identified four distinct molecular groups (A–D), each gene expression used as a conditioning input for image generation. The corresponding synthetic histology patches exhibited coherent tissue architectures and distinct morphological variations among clusters, consistent with real histological structures. Across all clusters, the generated images displayed high visual fidelity, realistic cellular organization, and smooth structural transitions comparable to real patches. Similar patterns were observed for the breast (TENX13), skin (MEND40), and lung (MEND90) centers (Fig.~\ref{fig:gen-TENX13}–\ref{fig:gen-MEND90}). In each center, local adaptation produced morphologically realistic and cluster-specific tissue structures. These results show that the locally adapted models generated organ-specific histology consistent with cluster-level gene expression patterns across multiple tissue types.

\subsection{Distribution alignment between synthetic and real histology features}

\begin{figure*}[htbp!]
    \centering
    \includegraphics[width=0.98\textwidth]{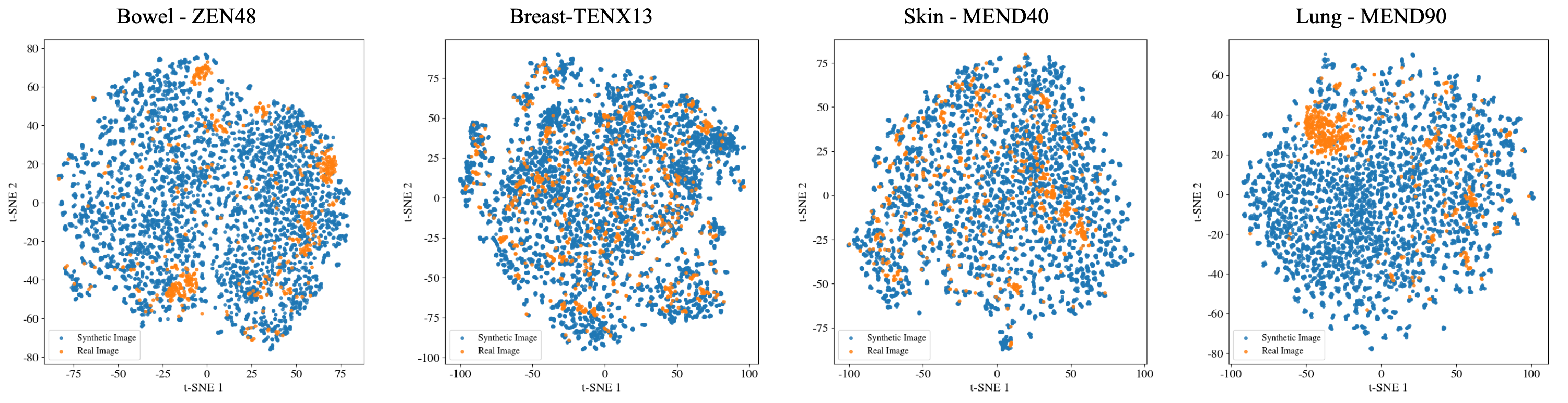}
    \caption{\textbf{Feature embedding similarity between real and synthetic histology patches across organs.} t-SNE visualization of feature embeddings extracted from the pretrained Conch pathology encoder for real (brown) and synthetic (blue) histology patches in four spatial transcriptomics (ST) centers: bowel (ZEN48), breast (TENX13), skin (MEND40), and lung (MEND90). In each organ, the synthetic patches exhibit distributions that closely overlap with those of the real samples while also extending the overall embedding space, indicating both high fidelity and enhanced diversity of the generated data. This embedding-level consistency demonstrates that the C2L-ST model effectively captures local tissue heterogeneity and morphological diversity across distinct anatomical domains.}
    \label{fig:emb}
\end{figure*}

\begin{figure*}[htbp!]
    \centering
    \includegraphics[width=0.98\textwidth]{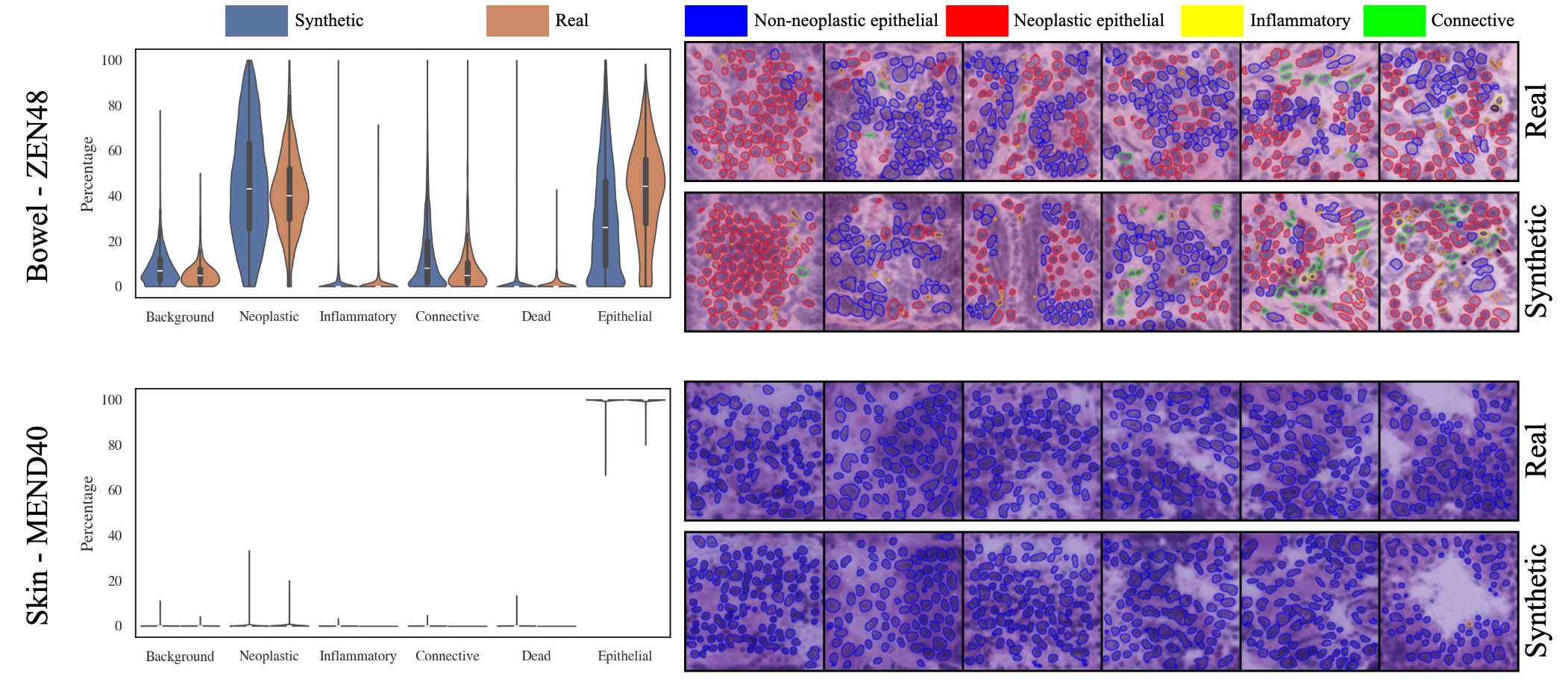}
    \caption{\textbf{Comparison of cellular composition and morphological fidelity between real and synthetic histology patches.} Violin plots on the left show the distribution of six major cell types, including background, neoplastic, inflammatory, connective, dead, and epithelial, quantified using HoverNet segmentation for real (orange) and synthetic (blue) images from bowel (ZEN48) and skin (MEND40) ST datasets. The segmentation visualizations on the right display corresponding real and synthetic tissue patches with color-coded cell-type annotations, where blue indicates non-neoplastic epithelial cells, red indicates neoplastic epithelial cells, yellow indicates inflammatory cells, and green indicates connective cells. The synthetic images reproduce cell-type distributions and spatial organization that closely resemble those of real samples, while also expanding the diversity of local cellular configurations. These results demonstrate that the C2L-ST framework preserves realistic cellular composition and structural integrity across distinct tissue domains.}
    \label{fig:cell}
\end{figure*}

\begin{table*}[htbp!]
\centering
\footnotesize
\setlength{\tabcolsep}{3pt}
\caption{\textbf{Comparison of cell-type composition between real and synthetic histology patches.}
Proportions (mean$\pm$SD) of six major cell types identified by HoverNet across four representative ST datasets.}
\label{tab:cell_composition}
\resizebox{\textwidth}{!}{%
\begin{tabular}{llcccccc}
\toprule
\textbf{Dataset (Organ)} & \textbf{Type} & \textbf{Background} & \textbf{Neoplastic} & \textbf{Inflammatory} & \textbf{Connective} & \textbf{Dead} & \textbf{Epithelial} \\
\midrule
\multirow{2}{*}{ZEN48 (Bowel)}  
& Real       & 0.060$\pm$0.066 & 0.411$\pm$0.176 & 0.010$\pm$0.049 & 0.085$\pm$0.118 & 0.013$\pm$0.042 & 0.421$\pm$0.209 \\
& Synthetic  & 0.086$\pm$0.079 & 0.449$\pm$0.247 & 0.015$\pm$0.053 & 0.133$\pm$0.154 & 0.017$\pm$0.060 & 0.299$\pm$0.240 \\
\midrule
\multirow{2}{*}{TENX13 (Breast)} 
& Real       & 0.007$\pm$0.019 & 0.972$\pm$0.071 & 0.003$\pm$0.018 & 0.004$\pm$0.025 & 0.001$\pm$0.010 & 0.014$\pm$0.052 \\
& Synthetic  & 0.009$\pm$0.029 & 0.944$\pm$0.134 & 0.007$\pm$0.036 & 0.011$\pm$0.057 & 0.004$\pm$0.033 & 0.025$\pm$0.089 \\
\midrule
\multirow{2}{*}{MEND40 (Skin)} 
& Real       & 0.0002$\pm$0.002 & 0.0026$\pm$0.015 & 0.0000$\pm$0.000 & 0.0000$\pm$0.000 & 0.0000$\pm$0.000 & 0.997$\pm$0.015 \\
& Synthetic  & 0.0002$\pm$0.003 & 0.0031$\pm$0.016 & 0.00002$\pm$0.001 & 0.00007$\pm$0.002 & 0.0002$\pm$0.003 & 0.996$\pm$0.017 \\
\midrule
\multirow{2}{*}{MEND90 (Lung)} 
& Real       & 0.183$\pm$0.204 & 0.701$\pm$0.267 & 0.006$\pm$0.068 & 0.072$\pm$0.196 & 0.023$\pm$0.103 & 0.015$\pm$0.056 \\
& Synthetic  & 0.199$\pm$0.232 & 0.624$\pm$0.320 & 0.010$\pm$0.072 & 0.127$\pm$0.255 & 0.023$\pm$0.099 & 0.018$\pm$0.071 \\
\bottomrule
\end{tabular}%
}
\end{table*}

We evaluated the distributional similarity between synthetic and real histology patches using feature embeddings extracted from the pretrained Conch pathology encoder~\cite{luVisuallanguageFoundationModel2024}, with implementation details provided in Methods~\ref{subsec:implement}.
Feature distributions were visualized with t-SNE for four representative ST centers (Fig.~\ref{fig:emb}).% : bowel (ZEN48), breast (TENX13), skin (MEND40), and lung (MEND90) 
Across all centers, synthetic samples generated by the locally adapted diffusion models overlapped with real tissue patches in the embedding space. Cosine similarity between real and synthetic embeddings further confirmed this alignment, with mean ($\pm$SD) values of 0.79$\pm$0.10 for ZEN48, 0.89$\pm$0.06 for TENX13, 0.87$\pm$0.05 for MEND40, and 0.81$\pm$0.07 for MEND90.
These high similarity scores indicate that the generated patches closely matched real tissue representations captured by the Conch encoder. t-SNE plots showed that the synthetic embeddings covered a much broader area within the shared feature space, reflecting an increased diversity of morphological patterns without evident domain separation from the real data. These results show that the locally adapted generative models achieved strong distributional alignment with real histology while maintaining feature diversity across multiple tissue types.

\subsection{Preservation of cellular composition in synthetic histology}

We further evaluated whether the generated histology patches preserved realistic cellular organization and composition. Cell segmentation and classification were performed using HoverNet~\cite{grahamHoVerNetSimultaneousSegmentation2019} to quantify six major types: background, neoplastic, inflammatory, connective, dead, and epithelial, across real and synthetic samples from four ST centers representing bowel (ZEN48), breast (TENX13), skin (MEND40), and lung (MEND90) tissues (Fig.~\ref{fig:cell}, Fig.~\ref{fig:cell_sup}, Table~\ref{tab:cell_composition}), with implementation details provided in Methods~\ref{subsec:implement}. Across all centers, the synthetic patches showed cell-type proportions consistent with those of real tissue. In ZEN48 (bowel), the dominant epithelial and neoplastic fractions were 0.30$\pm$0.24 and 0.45$\pm$0.25 in generated images, closely matching 0.42$\pm$0.21 and 0.41$\pm$0.18 in real tissue. For TENX13 (breast), neoplastic cells comprised nearly the entire tissue area, with proportions of 0.94$\pm$0.13 in synthetic and 0.97$\pm$0.07 in real samples. In MEND40 (skin), epithelial cells accounted for more than 99\% of both real and generated regions (0.997$\pm$0.015 vs 0.996$\pm$0.017). In MEND90 (lung), background, neoplastic, and connective components remained comparable between synthetic (0.20$\pm$0.23, 0.62$\pm$0.32, 0.13$\pm$0.26) and real (0.18$\pm$0.20, 0.70$\pm$0.27, 0.07$\pm$0.20) tissues. The accompanying segmentation overlays visualize the corresponding cell-level predictions, where nuclei and boundaries are color-coded by cell type. Synthetic patches reproduced the fine-grained spatial arrangement of epithelial and stromal cells observed in real tissues, maintaining realistic nuclear density and neighborhood organization. Overall, results confirmed that cell-type composition was well preserved in the generated histology patches. These results indicate that the locally adapted diffusion models produced biologically coherent cellular organization consistent with the tissue-specific characteristics of each center.

\begin{table*}[htbp!]
\centering
\footnotesize
\setlength{\tabcolsep}{3pt}
\caption{\textbf{Mean absolute error (MAE) of gene expression prediction with varying amounts of synthetic data.}
Each model was trained using either real data only, synthetic data only, or a combination of both. 
Synthetic-to-real ratios range from 0× (real only baseline) to 10× per real sample.}
\label{tab:mae_comparison}
\resizebox{\textwidth}{!}{%
\begin{tabular}{llccccccccccc}
\toprule
\multirow{2}{*}{\textbf{Dataset}} & \multirow{2}{*}{\textbf{Training setup}} & 
\multicolumn{11}{c}{\textbf{Synthetic samples per real sample}} \\
\cmidrule(lr){3-13}
 &  & Real only & 1× & 2× & 3× & 4× & 5× & 6× & 7× & 8× & 9× & 10× \\
\midrule
\multirow{2}{*}{ZEN48 (Bowel)} 
 & w/ Synthetic  & –      & 0.4078 & 0.4011 & 0.3972 & 0.3863 & 0.3819 & 0.3762 & 0.3692 & 0.3641 & 0.3627 & 0.3604 \\
 & w/ Real + Synthetic    & 0.4115 & 0.3993 & 0.3888 & 0.3789 & 0.3770 & 0.3699 & 0.3657 & 0.3607 & 0.3569 & 0.3539 & 0.3546 \\
\midrule
\multirow{2}{*}{TENX13 (Breast)}
 & w/ Synthetic  & –      & 0.3933 & 0.3804 & 0.3727 & 0.3674 & 0.3585 & 0.3526 & 0.3523 & 0.3458 & 0.3447 & 0.3419 \\
 & w/ Real + Synthetic    & 0.3903 & 0.3755 & 0.3636 & 0.3628 & 0.3559 & 0.3511 & 0.3463 & 0.3462 & 0.3405 & 0.3386 & 0.3391 \\
\midrule
\multirow{2}{*}{MEND40 (Skin)}
 & w/ Synthetic  & –      & 0.5492 & 0.5333 & 0.5241 & 0.5148 & 0.5084 & 0.5057 & 0.5193 & 0.5040 & 0.5033 & 0.5058 \\
 & w/ Real + Synthetic    & 0.5451 & 0.5298 & 0.5165 & 0.5116 & 0.5054 & 0.5000 & 0.4969 & 0.5074 & 0.4962 & 0.4990 & 0.5016 \\
\midrule
\multirow{2}{*}{MEND90 (Lung)}
 & w/ Synthetic  & –      & 0.4952 & 0.4936 & 0.4816 & 0.4770 & 0.4704 & 0.4629 & 0.4639 & 0.4559 & 0.4542 & 0.4524 \\
 & w/ Real + Synthetic    & 0.5030 & 0.4903 & 0.4879 & 0.4696 & 0.4670 & 0.4587 & 0.4558 & 0.4536 & 0.4488 & 0.4465 & 0.4469 \\
\bottomrule
\end{tabular}%
}
\end{table*}

\subsection{Synthetic image–gene pairs improve gene expression prediction}

\begin{figure*}[htbp!]
    \centering
    \includegraphics[width=0.98\textwidth]{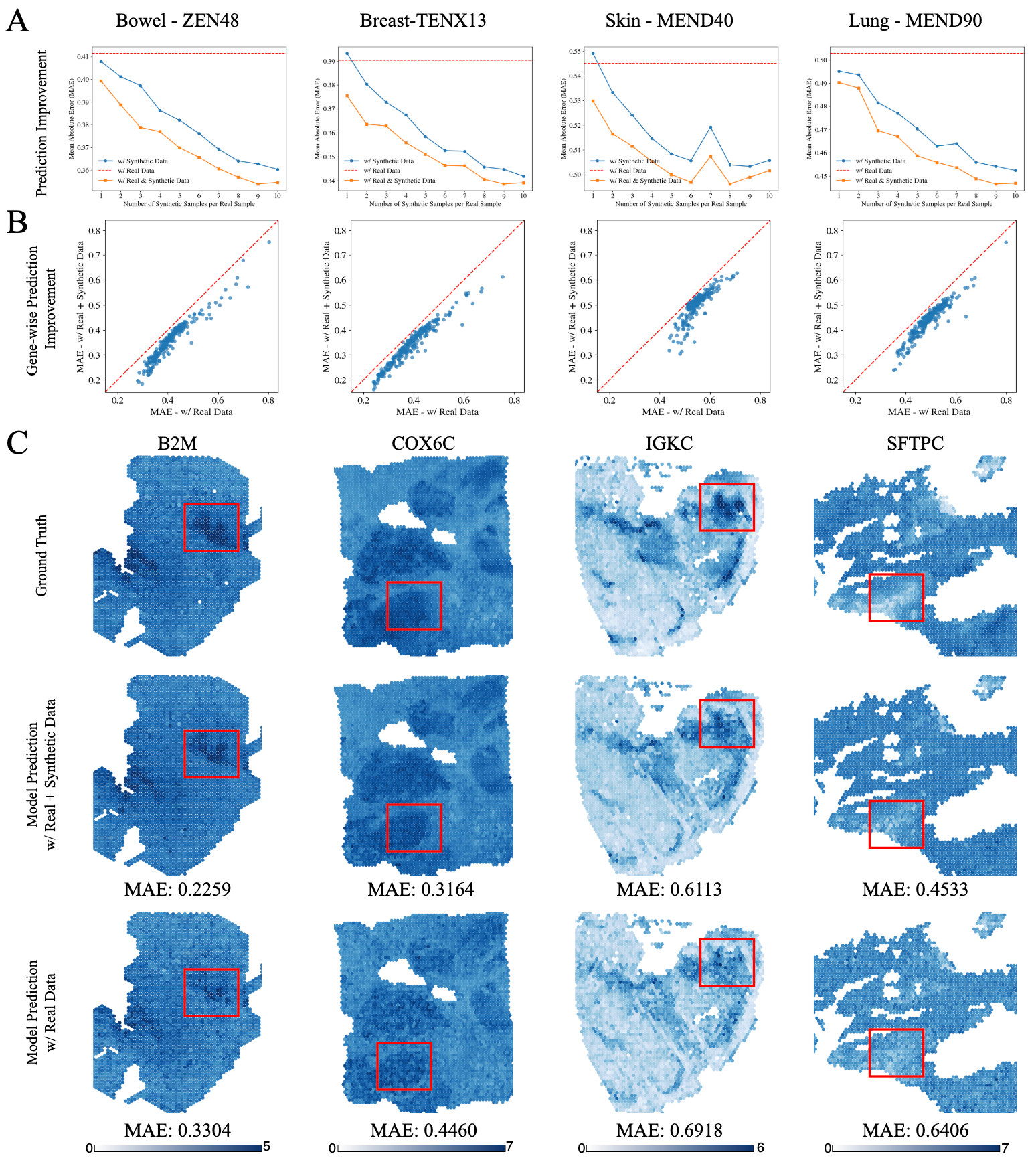}
    \caption{\textbf{Performance improvement in gene expression prediction enabled by the C2L-ST framework.} \textbf{(A)} Gene prediction performance evaluated across four local ST centers, including bowel (ZEN48), breast (TENX13), skin (MEND40), and lung (MEND90). Each plot shows the mean absolute error (MAE) as a function of the number of real samples per gene when training with real data only (orange) or with both real and synthetic data (blue). In all organs, incorporating synthetic samples leads to consistently lower prediction errors, particularly under data-limited conditions. All subsequent analyses, including panels (B) and (C), are based on models trained with real data combined with synthetic data at a synthetic-to-real ratio of 10$\times$. \textbf{(B)} Gene-wise comparison of prediction errors demonstrates that adding synthetic data reduces MAE for most genes, as indicated by the majority of points below the diagonal line. \textbf{(C)} Spatial expression maps of representative genes are shown for four organs: B2M in bowel, COX6C in breast, IGKC in skin, and SFTPC in lung. Each panel compares the ground truth, predictions with real and synthetic data, and predictions with real data only. The inclusion of synthetic data improves spatial coherence and preserves fine-grained molecular heterogeneity across distinct tissue domains.}
    \label{fig:improvement}
\end{figure*}

We evaluated the effect of synthetic data augmentation on gene expression prediction across four ST centers, with implementation details provided in Methods~\ref{subsec:implement}. % : bowel (ZEN48), breast (TENX13), skin (MEND40), and lung (MEND90)
Prediction accuracy was assessed by the MAE between predicted and observed gene expression values (Table~\ref{tab:mae_comparison}, Fig.~\ref{fig:improvement}). Across all datasets, models trained with both real and synthetic image–gene pairs achieved consistently lower MAE than those trained on real data alone. At a synthetic-to-real ratio of 10$\times$, the MAE decreased from 0.4115 to 0.3546 in ZEN48, from 0.3903 to 0.3391 in TENX13, from 0.5451 to 0.5016 in MEND40, and from 0.5030 to 0.4469 in MEND90. Averaged across genes, the inclusion of synthetic data (10$\times$) reduced the MAE by 0.0557$\pm$0.0227 for ZEN48, 0.0502$\pm$0.0217 for TENX13, 0.0473$\pm$0.0294 for MEND40, and 0.0529$\pm$0.0217 for MEND90, relative to real only training. Gene-wise comparisons (Fig.~\ref{fig:improvement}B) showed that the majority of genes exhibited lower prediction losses when synthetic data were included, with most data points positioned below the diagonal line across all organs, reflecting a consistent gene-level improvement rather than changes restricted to a subset of highly expressed genes.  Spatial prediction maps (Fig.~\ref{fig:improvement}C) further support these quantitative findings. For representative genes B2M (bowel), COX6C (breast), IGKC (skin), and SFTPC (lung), co-trained models reconstructed finer spatial gradients and smoother tissue boundaries compared with models trained on real data alone. These results confirm that incorporating synthetic image–gene pairs improves both quantitative accuracy and spatial coherence of gene expression inference across diverse tissue types.

\subsection{Sampling efficiency and scalability of local adaptation}

\begin{figure*}[htbp!]
    \centering
    \includegraphics[width=0.98\textwidth]{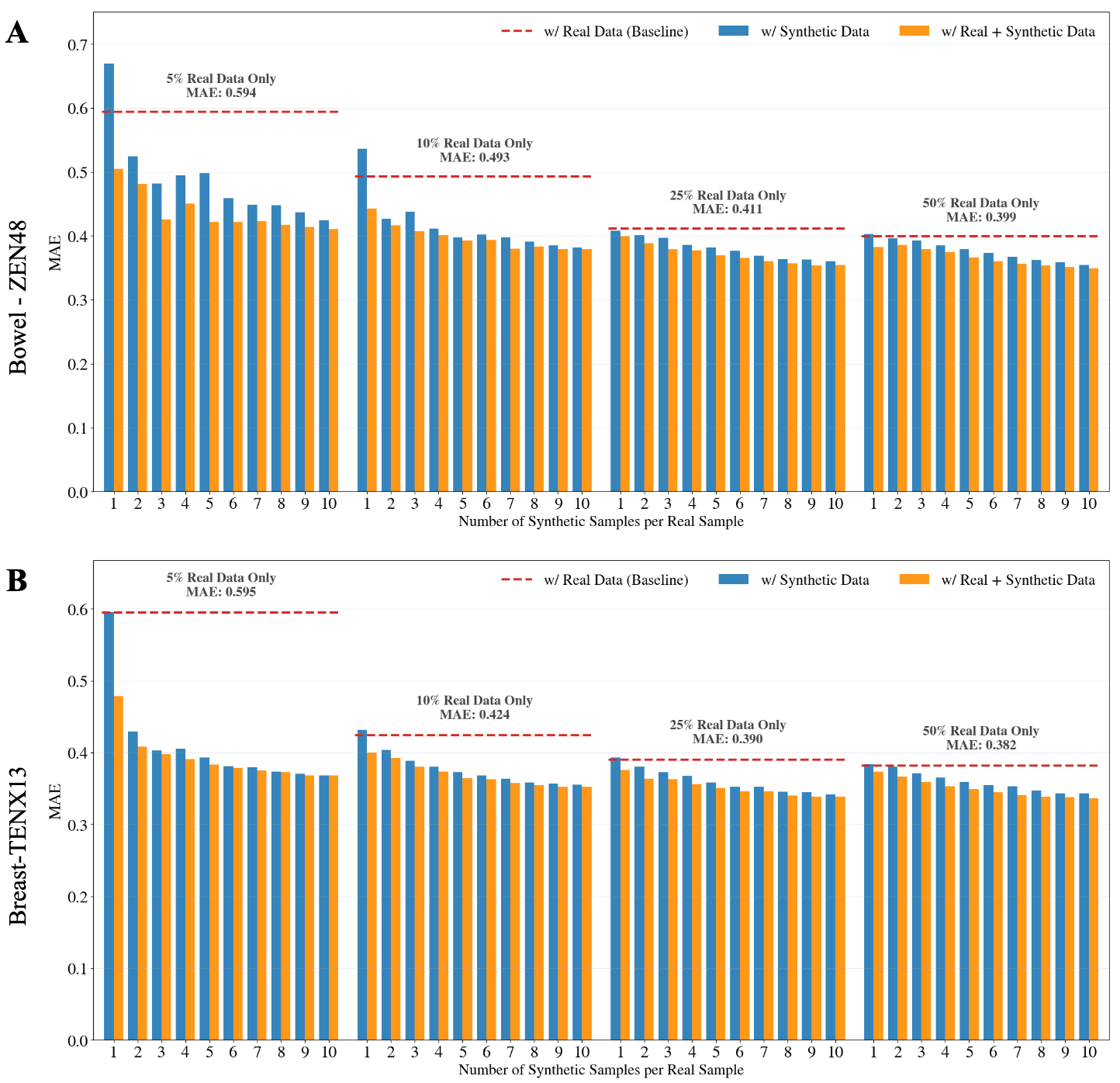}
    \caption{\textbf{Scaling analysis of gene expression prediction performance with varying synthetic-to-real data ratios.} \textbf{(A–B)} MAE of gene expression prediction for bowel (ZEN48) and breast (TENX13) datasets under different training configurations. Each group of bars represents models trained with synthetic data only (blue), real data only (red dashed line, baseline), and a combination of real and synthetic data (orange). Results are shown for four levels of real data availability, including 5\%, 10\%, 25\%, and 50\% of the full dataset, across synthetic-to-real ratios ranging from 1$\times$ to 10$\times$. In both datasets, adding synthetic data consistently reduces prediction error compared to models trained on real data alone, and performance continues to improve as the synthetic-to-real ratio increases.}
    \label{fig:ratio}
\end{figure*}

% We next evaluated the sampling efficiency of the proposed central–local framework by varying the proportion of ST spots used for local adaptation and subsequent synthetic data generation.
% Experiments were conducted on two representative centers, bowel (ZEN48) and breast (TENX13), using 5\%, 10\%, 25\%, and 50\% of the available spots from each slide (Fig.~\ref{fig:ratio}).

We next evaluated the sampling efficiency of the proposed central–local framework by varying the proportion of ST spots used for local adaptation and subsequent synthetic data generation.
To emulate a variety of data-limited conditions encountered in practical ST experiments, we conducted experiments in two representative centers, bowel (ZEN48) and breast (TENX13), using 5\%, 10\%, 25\%, and 50\% of the available spots from each slide (Fig.~\ref{fig:ratio}). 
The selected subsets were used for gene-conditioned local adaptation of the central model, followed by the generation of synthetic image–gene pairs that were combined with real data for model co-training. Across all sampling configurations, incorporating synthetic data consistently reduced prediction loss compared with real-only baselines. For ZEN48, when only 5\% of real spots were used, the model trained solely on real data achieved an MAE of 0.5944, whereas co-training with synthetic data reduced the MAE to 0.4104. At higher sampling ratios, the corresponding MAE values decreased from 0.4928 to 0.3792 (10\%), from 0.4115 to 0.3546 (25\%), and from 0.3994 to 0.3490 (50\%).
Similarly, for TENX13, co-training with synthetic data improved the MAE from 0.5950 to 0.3680 (5\%), from 0.4240 to 0.3523 (10\%), from 0.3903 to 0.3391 (25\%), and from 0.3819 to 0.3369 (50\%). These results demonstrate that the inclusion of synthetic data recovered most of the predictive performance even when only 5–10\% of real spots were used for local adaptation, and that the benefit gradually saturated beyond 25\%. These findings indicate that the central–local framework achieves high sampling efficiency and scalability, maintaining robust predictive accuracy under limited data conditions while requiring only a small fraction of real spatial samples for effective adaptation.

\section{Discussion}\label{sec12}
%%%%%%%%%%%%%%%%%%%%%%%%
ST enables the spatial mapping of gene expression within intact tissue contexts, providing essential insights into cellular organization and molecular interactions. Despite its transformative potential in decoding spatial heterogeneity and disease mechanisms, the application of ST remains constrained by limited data availability and insufficient spatial resolution~\cite{choeAdvancesChallengesSpatial2023,smithChallengesOpportunitiesClinical2024}. High experimental costs, complex protocols, and data privacy concerns across institutions restrict large-scale data sharing and collaborative analysis. Even when data are accessible, variations among centers and sequencing platforms often result in large domain gaps that hinder model generalization~\cite{attaComputationalChallengesOpportunities2021}. These challenges collectively limit the training of robust generative and predictive models and impede the broader biomedical adoption of ST technologies.

These limitations have constrained the development of computational models for ST~\cite{attaComputationalChallengesOpportunities2021}. Existing approaches, such as ST-Net~\cite{pangLeveragingInformationSpatial2021} and EGN~\cite{yangExemplarGuidedDeep2023} have demonstrated the feasibility of predicting gene expression from histology, yet their progress has been slowed by the scarcity of available data. The restricted data scale prevents these models from capturing the complex relationships between tissue morphology and molecular states, limiting their generalization across biological and technical contexts. Generative modeling provides a promising direction to alleviate these challenges, as it enables the synthesis of realistic tissue images and molecular patterns without directly relying on large quantities of sensitive clinical data~\cite{yellapragadaPixCellGenerativeFoundation2025,nicholImprovedDenoisingDiffusion2021}. Building upon this concept, we developed C2L-ST, a central-to-local generative framework designed to integrate large-scale morphological knowledge with limited molecular supervision. The framework leverages abundant unlabeled pathology images to learn broad tissue representations and adapts them to local molecular contexts through gene-conditioned fine-tuning. This design not only mitigates data scarcity and enhances cross-center usability but also provides synthetic data that significantly improves downstream tasks such as gene expression prediction. Through scalable and locally adaptive learning, C2L-ST offers a practical pathway toward more generalizable molecular inference in ST.

The success of generative modeling in ST ultimately depends on the ability to incorporate molecular-level information with sufficient precision. While many pathology-based frameworks can synthesize realistic histological patterns, most lack direct molecular supervision. They are typically trained with categorical annotations such as cell type labels or binary indicators of gene expression, which capture only coarse biological features rather than quantitative molecular variation~\cite{niehuesUsingHistopathologyLatent2024,liToPoFMTopologyGuidedPathology2025,ohDiffMixDiffusionModelbased2023, louMultimodalDenoisingDiffusion2024}. As a result, these models can reproduce general tissue morphology but fail to represent the continuous relationships between gene expression and structural organization, limiting their suitability for ST. RNA-CDM introduces molecular conditioning through gene expression profiles, yet it relies on bulk RNA-seq data that lack spatial specificity and do not match the patch-level resolution required for ST~\cite{carrillo-perezGenerationSyntheticWholeslide2025a}. By contrast, our C2L-ST incorporates molecular-level conditioning at the patch level, using continuous spatial gene expression values to guide image generation. This design allows the model to capture fine molecular gradients and localized transcriptional diversity, achieving spatially resolved morphology–molecule correspondences that meet the resolution required for accurate modeling in ST.

Our C2L-ST framework operates through a two-stage learning process that combines large-scale morphological pretraining with targeted molecular adaptation. In the first stage, the central model learns broad histological priors from extensive pathology image collections, capturing diverse tissue architectures and morphological variability. During the second stage, local adaptation with limited ST data aligns these morphological representations with molecular information specific to each tissue domain. The resulting synthetic images exhibit high visual fidelity (Fig.~\ref{fig:gen-ZEN48}, Fig.~\ref{fig:gen-TENX13}–\ref{fig:gen-MEND90}), with realistic cellular organization and coherent structural patterns that closely resemble real tissues  (Fig.~\ref{fig:cell}, Fig.~\ref{fig:cell_sup}, Table~\ref{tab:cell_composition}). Quantitative analyses further show that synthetic and real samples share overlapping embedding distributions, while the generated data maintain broader and more diverse feature ranges (Fig.~\ref{fig:emb}), indicating both high fidelity and preserved variability. Even when randomly dropping five genes from the conditioning input, a scenario that mimics the common dropout phenomenon observed during ST experiments~\cite{wangSprodDenoisingSpatial2022,molladestaAdvancementsSinglecellRNA2025}, the model still produces visually consistent and biologically plausible synthetic images (Fig.~\ref{fig:gene_dropout}), highlighting its robustness to incomplete molecular supervision. Notably, the model achieves this consistency even under highly data-limited conditions, demonstrating strong adaptability and effective use of scarce molecular supervision. These characteristics were consistently observed across multiple organs, demonstrating the robustness and generalizability of the framework. Moreover, clusters derived from gene expression features correspond closely with morphological clusters in the generated images (Fig.~\ref{fig:gen-ZEN48}, Fig.~\ref{fig:gen-TENX13}–\ref{fig:gen-MEND90}), confirming that the model effectively captures the relationship between molecular signals and tissue morphology rather than merely reproducing visual structures. 

Beyond its generative capability, C2L-ST provides tangible benefits for downstream computational and biological applications. The synthetic data produced by the framework significantly enhances gene expression prediction (Fig.~\ref{fig:improvement}, Table~\ref{tab:mae_comparison}), demonstrating that molecularly informed image generation can directly strengthen spatial inference tasks, as supported by quantitative comparisons. Although numerous models have been developed to predict molecular or categorical features from histology, their performance remains constrained by limited data availability and domain discrepancies across centers. Unlike previous generative frameworks that focus primarily on producing images for classification-level tasks~\cite{yellapragadaPathLDMTextConditioned2023}, C2L-ST operates at a molecular resolution, conditioning on continuous gene expression profiles. This shift could improve predictive modeling, including enhanced patch-level prediction accuracy (Fig.~\ref{fig:classification_improvement}), and enable the exploration of morphological variations under controlled molecular conditions, providing a potential framework for investigating the relationship between gene regulation and tissue phenotype, similar to recent generative approaches that model the morphological consequences of transcriptional perturbations~\cite{wangPredictionCellularMorphology2025}. These existing approaches stand to benefit from C2L-ST, which can serve as a source of high-quality, domain-adapted synthetic data to expand and harmonize training resources. Importantly, the framework achieves stable performance with only small amounts of local data (Fig.~\ref{fig:ratio}), making it highly data-efficient and cost-effective. This property provides a practical solution to the sparse sampling inherent in current ST protocols and can be further combined with emerging smart region-of-interest selection strategies (S2Omics)~\cite{yuanDesigningSmartSpatial2025} and sparse sampling optimization methods (S2S-ST)~\cite{fangSparser2SparseSingleshotSparsertoSparse2025} to improve data acquisition efficiency. These advances help mitigate the limitations imposed by sequencing only selected tissue regions, enabling scalable and resource-efficient ST analysis.

The conditioning strategy of the current framework provides a direct and effective means of incorporating molecular information, but it potentially remains relatively coarse in molecular resolution and may omit subtle or low-abundance features, which is an important direction for future studies. Future refinements could expand the conditioning space to include additional biological signals such as less expressed genes, alternative molecular markers, or complementary spatial modalities. Incorporating single-cell references, cell segmentation maps, and graph-based representations of cellular topology could further enrich the description of cell identity, spatial organization, and intercellular communication~\cite{anderssonSinglecellSpatialTranscriptomics2020,wanIntegratingSpatialSinglecell2023,lohoffIntegrationSpatialSinglecell2022}. These extensions would allow the model to integrate multi-level biological context and capture complex molecular–structural interactions with higher precision. Moreover, the same principle could be extended to other spatial omics modalities, such as spatial proteomics~\cite{wuROSIEAIGeneration2025}, providing a potential pathway toward unified molecular–morphological modeling across diverse datasets. Collectively, these advances in conditioning design and spatial modeling would enhance the interpretability, robustness, and general applicability of the C2L-ST framework.

In summary, the C2L-ST framework addresses fundamental challenges in ST by combining large-scale morphological learning with targeted molecular adaptation. By leveraging abundant pathology images without molecular labels and adapting to limited local data, the framework effectively mitigates data scarcity, institutional data-sharing constraints, and spatial resolution limitations that have long restricted the field. The generated synthetic data faithfully reproduces both morphological and molecular diversity, improves downstream gene expression prediction, and provides new opportunities for multi-center collaboration under secure data governance. Beyond the current study, the principles established by C2L-ST suggest a scalable paradigm for integrating molecular information into digital pathology and for developing generative models that connect the histological and genomic dimensions of tissue organization. As spatially resolved technologies continue to evolve, such frameworks will be essential for translating complex molecular landscapes into interpretable and actionable biological insights.

\section{Data availability}

All pathology image data used for training were obtained from the HistAI collection~\cite{nechaevSPIDERComprehensiveMultiOrgan2025}, which is publicly available at \href{https://huggingface.co/collections/histai/spider-models-and-datasets}{https://huggingface.co/collections/histai/spider-models-and-datasets}. All ST datasets used in this study were derived from the HEST-1K dataset~\cite{jaumeHEST1kDatasetSpatial2024}, accessible through \href{https://huggingface.co/datasets/MahmoodLab/hest}{https://huggingface.co/datasets/MahmoodLab/hest}.

\backmatter

\bibliography{sn-bibliography}% common bib file
%% if required, the content of .bbl file can be included here once bbl is generated
%%\input sn-article.bbl

\clearpage
\begin{appendices}

\section{Extended data}\label{secA1}

\begin{figure*}[htbp!]
    \centering
    \includegraphics[width=0.7\textwidth]{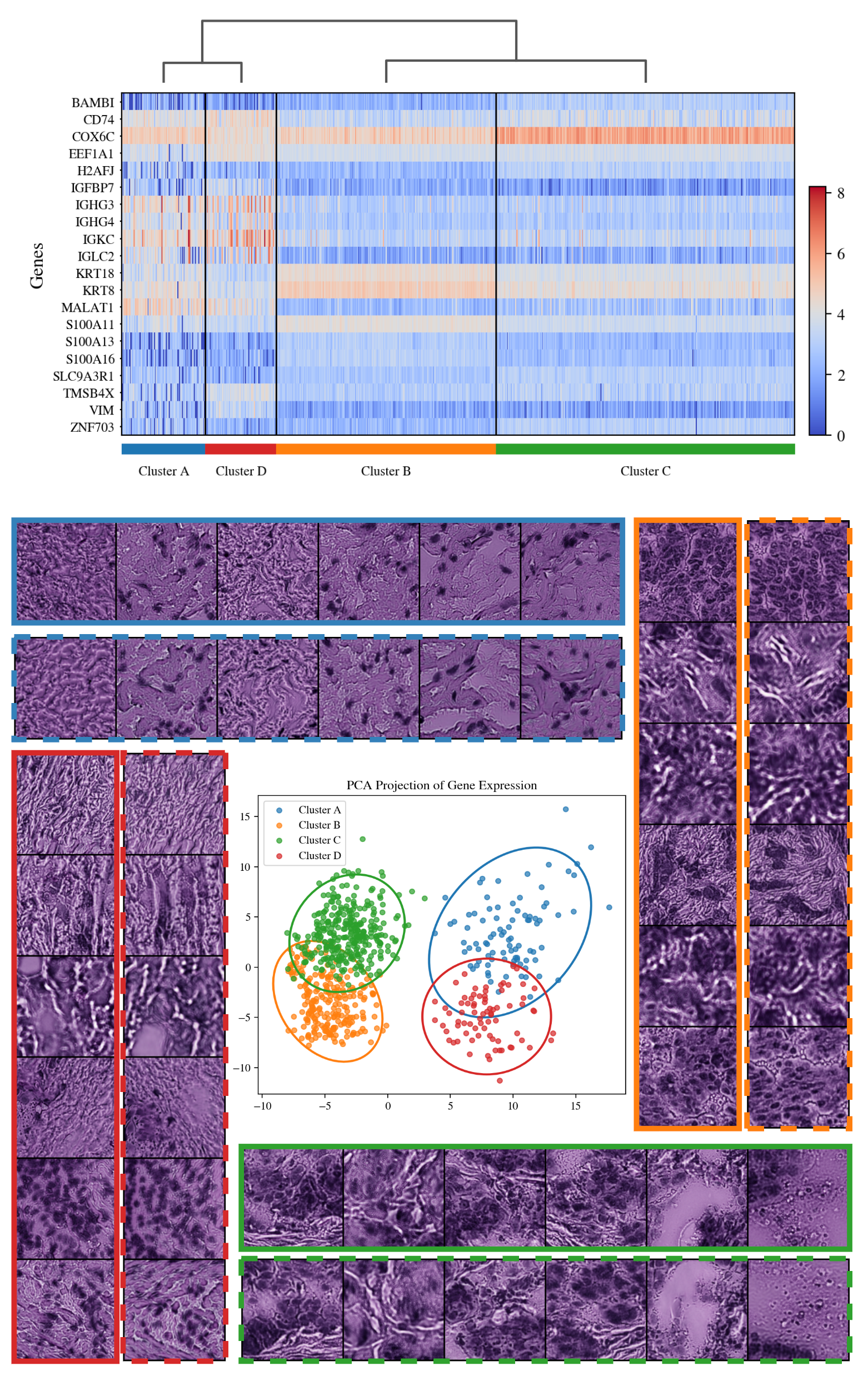}
    \caption{\textbf{Molecular and morphological correspondence in generated histology patches for the breast (TENX13) dataset.} 
    The heatmap shows hierarchical clustering of representative genes, revealing four distinct molecular clusters (A–D). Each cluster is associated with characteristic histology patches from synthetic images, shown below with color-coded borders corresponding to the cluster identity. Dashed borders indicate synthetic images, and solid borders indicate real images. The PCA projection of gene expression highlights the separation among clusters in molecular space. Representative synthetic patches within each cluster display distinct structures consistent with their molecular signatures, indicating that the model captures the relationship between transcriptional patterns and local morphology.}
    \label{fig:gen-TENX13}
\end{figure*}

\begin{figure*}[htbp!]
    \centering
    \includegraphics[width=0.7\textwidth]{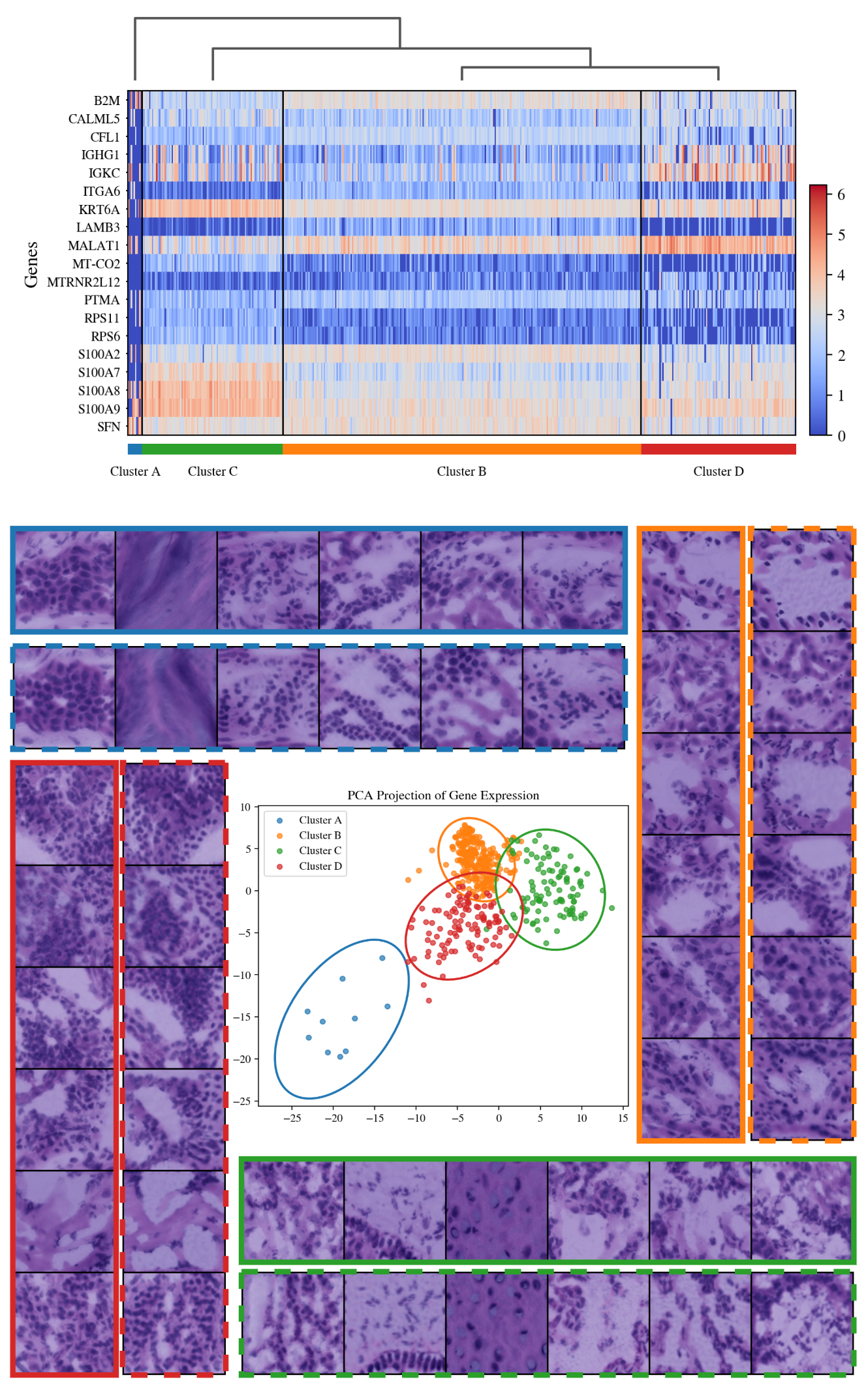}
    \caption{\textbf{Molecular and morphological correspondence in generated histology patches for the skin (MEND40) dataset.} The heatmap shows hierarchical clustering of representative genes, revealing four distinct molecular clusters (A–D). Each cluster is associated with characteristic histology patches from synthetic images, shown below with color-coded borders corresponding to the cluster identity. Dashed borders indicate synthetic images, and solid borders indicate real images. The PCA projection of gene expression highlights the separation among clusters in molecular space. Representative synthetic patches within each cluster display distinct morphologies consistent with their molecular signatures, indicating that the model captures the relationship between transcriptional patterns and local morphology.}
    \label{fig:gen-MEND40}
\end{figure*}

\begin{figure*}[htbp!]
    \centering
    \includegraphics[width=0.7\textwidth]{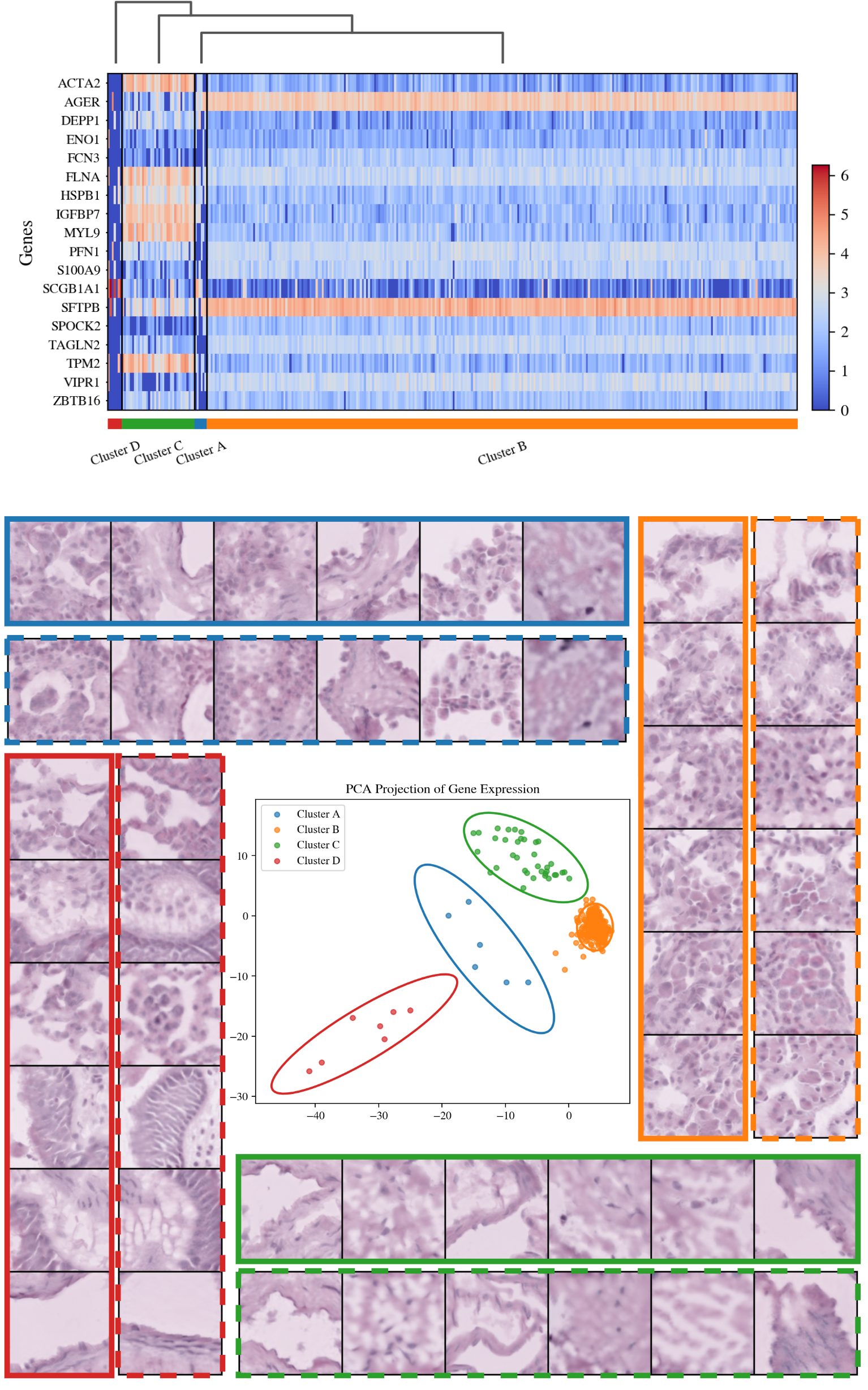}
    \caption{\textbf{Molecular and morphological correspondence in generated histology patches for the lung (MEND90) dataset.} The heatmap shows hierarchical clustering of representative genes, revealing four distinct molecular clusters (A–D). Each cluster is associated with characteristic histology patches from synthetic images, shown below with color-coded borders corresponding to the cluster identity. Dashed borders indicate synthetic images, and solid borders indicate real images. The PCA projection of gene expression highlights the separation among clusters in molecular space. Representative synthetic patches within each cluster display distinct structures consistent with their molecular signatures, indicating that the model captures the relationship between transcriptional patterns and local morphology.}
    \label{fig:gen-MEND90}
\end{figure*}

\begin{figure*}[htbp!]
    \centering
    \includegraphics[width=0.95\textwidth]{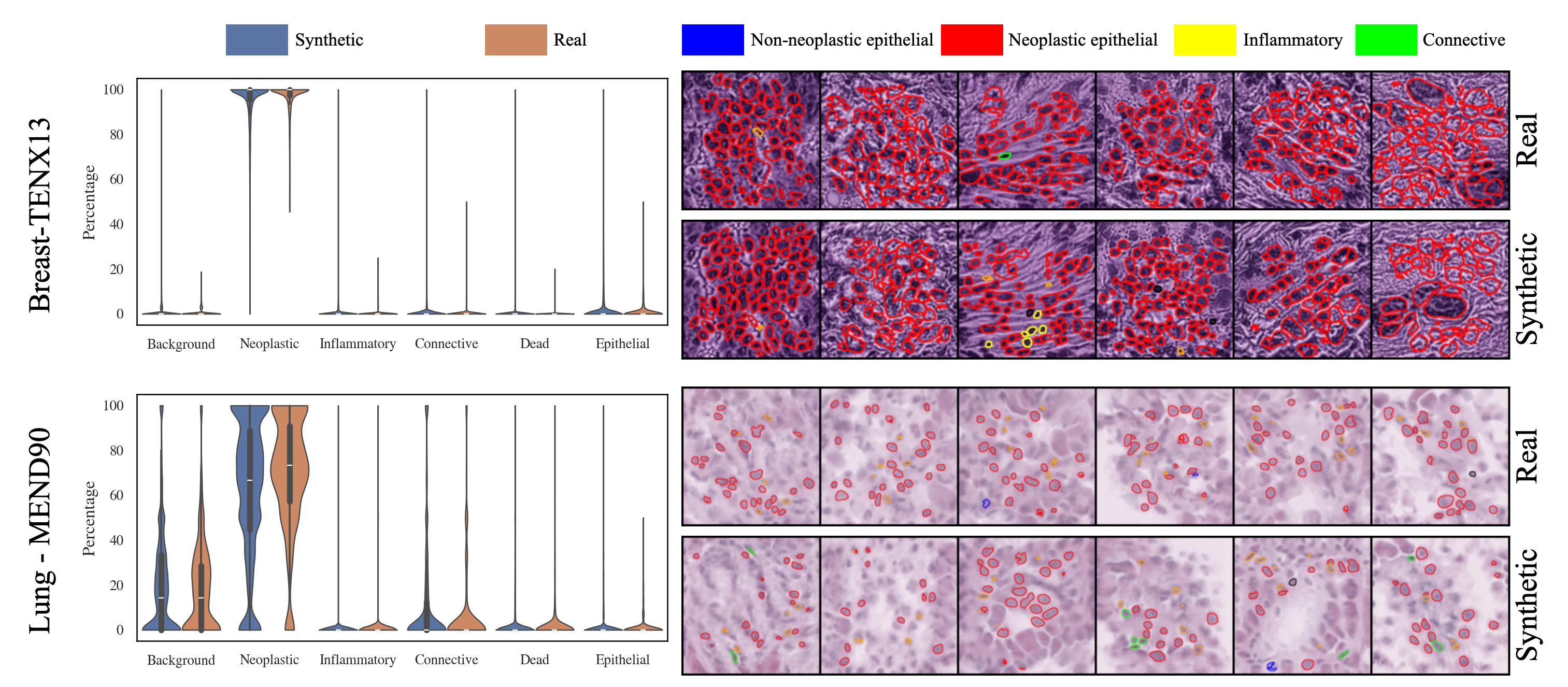}
    \caption{\textbf{Comparison of cellular composition and morphological fidelity between real and synthetic histology patches in additional organs.} 
    Violin plots on the left show the distribution of six major cell types, including background, neoplastic, inflammatory, connective, dead, and epithelial, quantified using HoverNet segmentation for real (orange) and synthetic (blue) images from breast (TENX13) and lung (MEND90) ST datasets. The segmentation visualizations on the right display corresponding real and synthetic tissue patches with color-coded cell-type annotations, where blue indicates non-neoplastic epithelial cells, red indicates neoplastic epithelial cells, yellow indicates inflammatory cells, and green indicates connective cells. The synthetic images reproduce cell-type distributions and spatial organization that closely resemble those of real samples, while also expanding the diversity of local cellular configurations. These results demonstrate that the C2L-ST framework preserves realistic cellular composition and structural integrity across distinct tissue domains.}
    \label{fig:cell_sup}
\end{figure*}

\begin{figure*}[htbp!]
    \centering
    \includegraphics[width=0.95\textwidth]{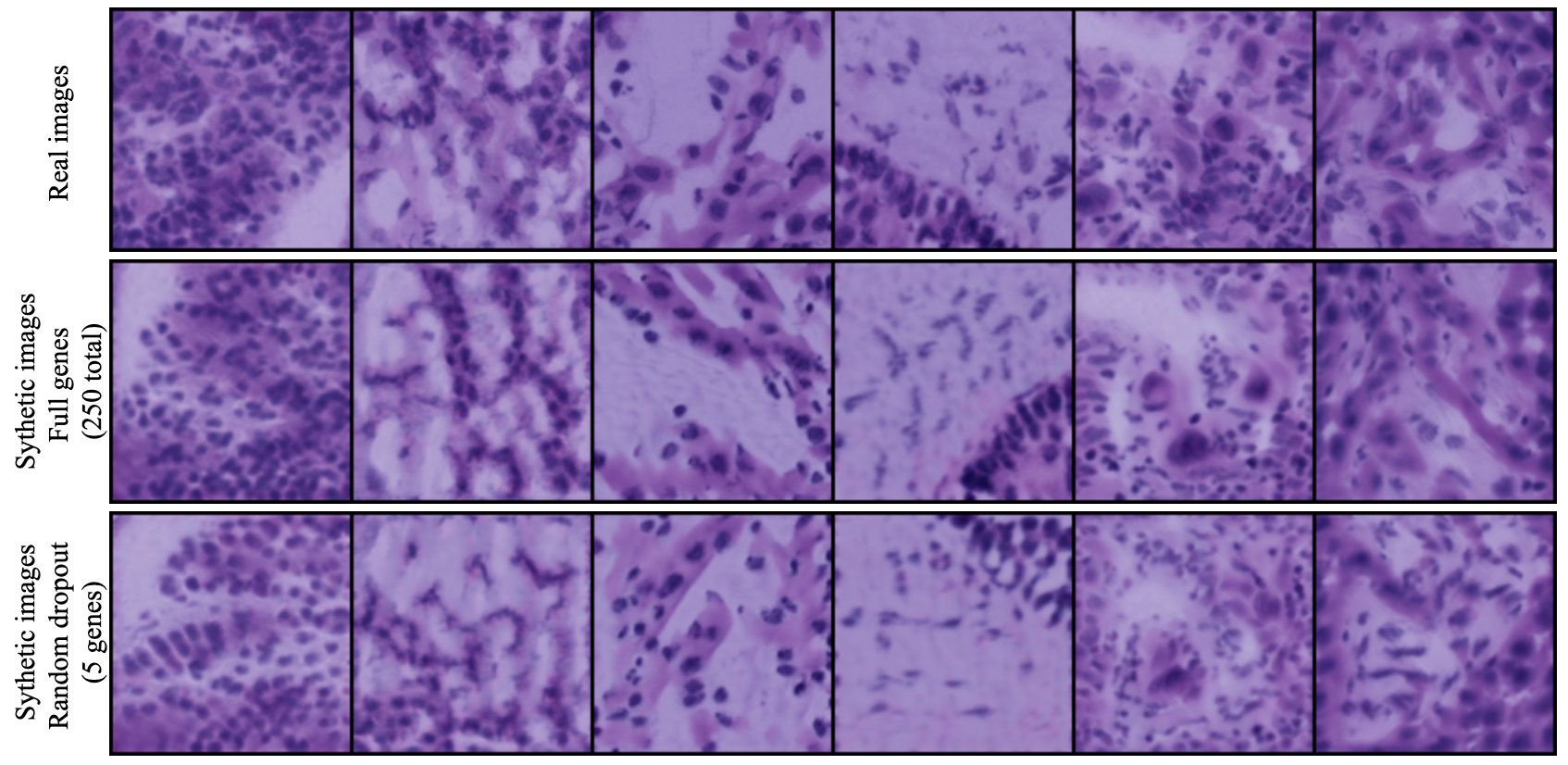}
    \caption{\textbf{Robustness of C2L-ST to gene dropout in conditioning inputs on the MEND40 (skin) dataset.} Representative real histology patches and synthetic images generated under two conditions are shown. The first synthetic set is conditioned on the complete set of 250 genes, while the second is generated after randomly dropping five genes from the conditioning input to mimic the common dropout phenomenon observed during ST experiments. The generated images remain visually consistent with real tissues, preserving cellular organization and structural integrity, which demonstrates the robustness of the C2L-ST framework to incomplete molecular supervision.}
    \label{fig:gene_dropout}
\end{figure*}

\begin{figure*}[htbp!]
    \centering
    \includegraphics[width=0.95\textwidth]{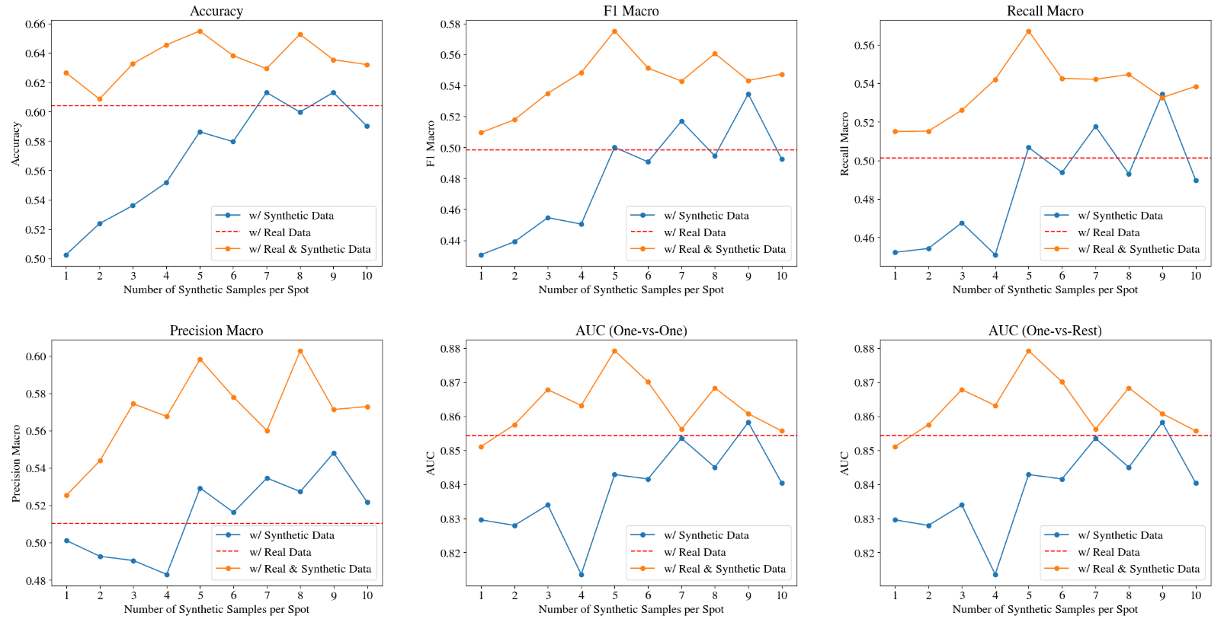}
    \caption{\textbf{Evaluation of synthetic data for improving class-level prediction performance on the ZEN48 (bowel) dataset.} Classification models were trained with synthetic data only (blue), real data only (red dashed line, baseline), and a combination of real and synthetic data (orange). Performance was assessed using accuracy, F1 macro, recall macro, precision macro, and area under the curve (AUC) for both one-vs-one and one-vs-rest configurations. Each x-axis represents the number of synthetic samples per real sample, corresponding to synthetic-to-real ratios ranging from 1$\times$ to 10$\times$. Across all evaluated metrics, incorporating synthetic data improved classification performance compared with models trained on real data alone, demonstrating that molecularly guided synthetic histology generated by the C2L-ST framework can enhance not only molecular prediction but also class-level discriminative tasks.}
    \label{fig:classification_improvement}
\end{figure*}

\end{appendices}

\end{document}